\documentclass{article}

\usepackage{microtype}
\usepackage{graphicx}
\usepackage{caption}
\usepackage{subcaption}
\usepackage{booktabs} 

\usepackage{hyperref}
\usepackage{url}

\usepackage[accepted]{icml2024}

\usepackage{bm}
\usepackage{amsmath}
\usepackage{amssymb}
\usepackage{mathtools}
\usepackage{xcolor}
\usepackage{amsthm}
\usepackage{nicefrac}       
\usepackage{pifont}
\usepackage{thmtools, thm-restate}
\usepackage{xspace}
\usepackage{multirow}

\usepackage[capitalize,noabbrev]{cleveref}

\input{defs.tex}
\newcommand{\methodname}{\textsc{PCLaSt}\xspace}

\icmltitlerunning{PcLast: Discovering Plannable Continuous Latent States}

\begin{document}

\twocolumn[
\icmltitle{PcLast: Discovering Plannable Continuous Latent States}



\icmlsetsymbol{equal}{*}

\begin{icmlauthorlist}
\icmlauthor{Anurag Koul}{equal,msr}
\icmlauthor{Shivakanth Sujit}{equal,intern,mila,ets}
\icmlauthor{Shaoru Chen}{msr}
\icmlauthor{Ben Evans}{nyu}
\icmlauthor{Lili Wu}{msr}
\icmlauthor{Byron Xu}{msr}
\icmlauthor{Rajan Chari}{msr}
\icmlauthor{Riashat Islam}{mila,mcgill}
\icmlauthor{Raihan Seraj}{mila,mcgill}
\icmlauthor{Yonathan Efroni}{meta}
\icmlauthor{Lekan Molu}{msr}
\icmlauthor{Miro Dudik}{msr}
\icmlauthor{John Langford}{msr}
\icmlauthor{Alex Lamb}{msr}
\end{icmlauthorlist}

\icmlaffiliation{msr}{Microsoft Research}
\icmlaffiliation{meta}{Meta}
\icmlaffiliation{nyu}{New York University}
\icmlaffiliation{mila}{Mila - Quebec AI Institute}
\icmlaffiliation{ets}{ETS Montreal}
\icmlaffiliation{mcgill}{McGill University}
\icmlaffiliation{intern}{Work done as Intern at Microsoft, NYC}

\icmlcorrespondingauthor{Anurag Koul}{anuragkoul@microsoft.com}
\icmlcorrespondingauthor{Shivakanth Sujit}{shivakanth.sujit.1@ens.etsmtl.ca}
\icmlcorrespondingauthor{Alex Lamb}{lambalex@microsoft.com}

\icmlkeywords{Representation Learning, Latent State, Control, Goal conditioned planning}

\vskip 0.3in
]



\printAffiliationsAndNotice{\icmlEqualContribution} 

\begin{abstract}
Goal-conditioned planning benefits from learned low-dimensional representations of rich observations. While compact latent representations typically learned from variational autoencoders or inverse dynamics enable goal-conditioned decision making, they ignore state reachability, hampering their performance. In this paper, we learn a representation that associates reachable states together for effective planning and goal-conditioned policy learning. We first learn a latent representation with multi-step inverse dynamics (to remove distracting information), and then transform this representation to associate reachable states together in $\ell_2$ space. Our proposals are rigorously tested in various simulation testbeds. Numerical results in reward-based settings show significant improvements in sampling efficiency. Further, in reward-free settings this approach yields layered state abstractions that enable computationally efficient hierarchical planning for reaching ad hoc goals with zero additional samples.

\end{abstract}

\section{Introduction}
\label{sec:introduction}

Deep reinforcement learning (RL) has emerged as a choice tool in mapping rich and complex perceptual information to compact low-dimensional representations for onward (motor) control in virtual environments~\cite{silver2016mastering}, software simulations~\cite{brockman2016openai}, and hardware-in-the-loop tests~\cite{finn2017deep}. Its impact traverses diverse disciplines spanning games \citep{moravvcik2017deepstack, brown2018superhuman}, virtual control~\citep{tunyasuvunakool2020dm_control}, healthcare~\citep{johnson2016mimic}, and autonomous driving~\citep{maddern20171, yu2018bdd100k}. Fundamental catalysts that have spurred these advancements include progress in algorithmic innovations~\citep{mnih2013playing, hessel2017rainbow, schrittwieser2020mastering} and learned (compact) latent representations \citep{bellemare2019geometric,lyle2021effect,lan2022generalization, rueckert2023cr,lan2023bootstrapped}.

\begin{figure}[b!]
\centering
\begin{subfigure}{.15\textwidth}
  \centering
  \includegraphics[scale=0.25]{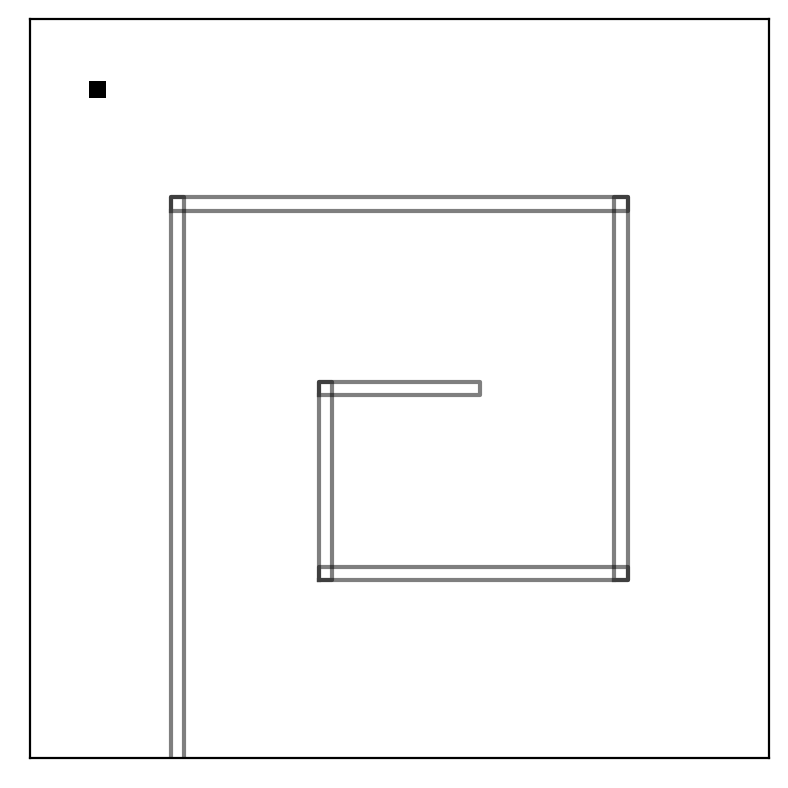}
  \caption{Maze-Spiral}
  \label{fig:maze-spiral-env}
\end{subfigure}%
\begin{subfigure}{.15\textwidth}
  \centering
    \includegraphics[scale=0.25]{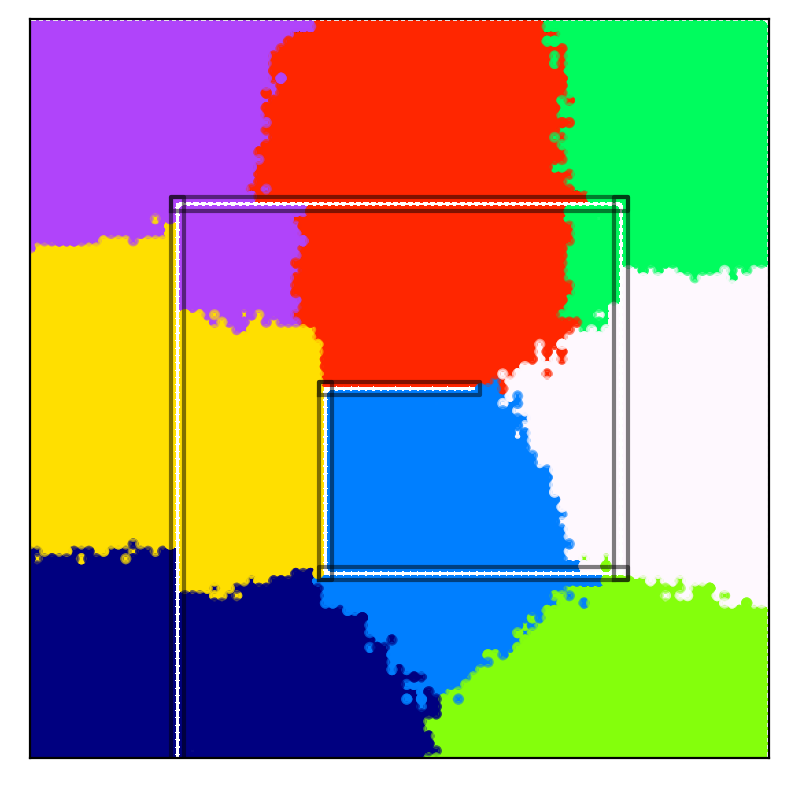} 
    \caption{ACRO}
  \label{fig:overview-acro-clustering}
\end{subfigure}
\begin{subfigure}{.15\textwidth}
  \centering
    \includegraphics[scale=0.25]{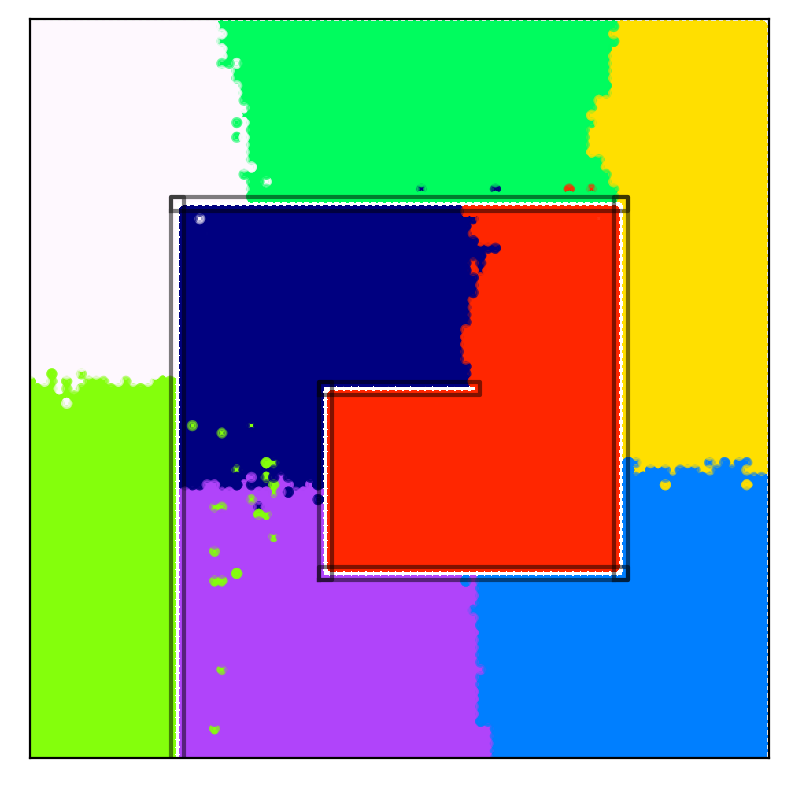}
    \caption{\methodname}
  \label{fig:pclast-clustering}
\end{subfigure}
\caption{Comparative view of clustering representations learned for a 2D maze environment with spiral walls (a). The agent's location is marked by black-dot in the maze image. The clustering of representations learned via ACRO (b) and \methodname (c) are overlaid on the maze image.}
\label{fig:overview-embedding-clustering}
\end{figure}

Latent representations, typically learned by variational autoencoders \citep{kingma2013auto} or inverse dynamics \citep{paster2020planning,wu2023agentcentric}, are mappings from high-dimensional observation spaces to a reduced space of essential information where extraneous perceptual information has already been discarded. Good compact representations foster sample efficiency in learning-based control settings \citep{ha2018world, lamb2022guaranteed}. Latent representations however often fail to correctly model the underlying states' affordances. Consider an agent in the 2D maze of \cref{fig:maze-spiral-env}. If we learn a typical representation (such as ACRO of \citealp{islam2022agent}) and cluster it, we observe it correctly identifies the agent's (low-level) position information as indicated by states with nearby coordinate position falling in the same clusters. However, it ignores the scene geometry such as the wall barriers so that states naturally demarcated by obstacles are clustered together (see in \cref{fig:overview-acro-clustering}). This inadequacy in creating poor abstractions is a drag on the efficacy of planning and deep RL algorithms despite their impressive showings in the last few years.

\begin{figure*}[tb!]
    \centering
    \includegraphics[scale=0.97]{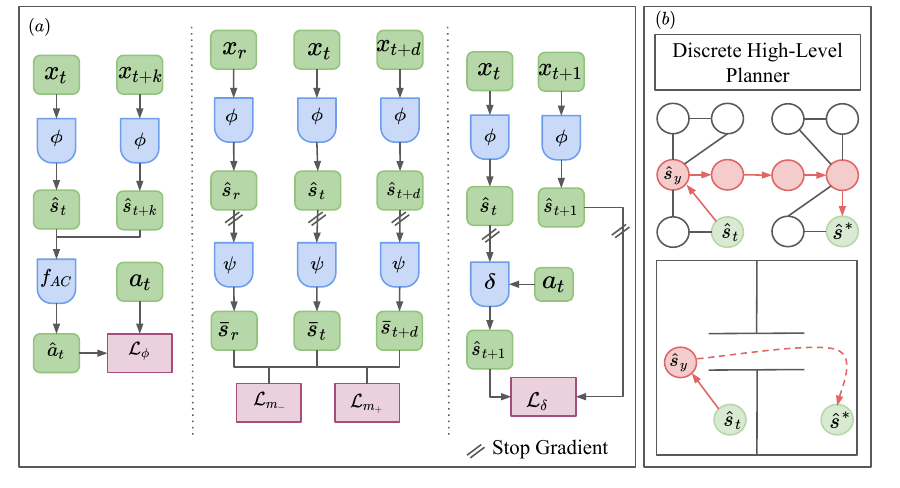}
    \caption{(a) Overview of the proposed method: (a) The encoder $\phi$, which maps observations $x$ to continuous latent states $\hat{s}$, is learned with a multi-step inverse model $f_{\mathrm{AC}}$ (left). A temporal contrastive objective ($\mathcal{L}_{m_-}$ and $\mathcal{L}_{m_+}$) is used to learn a metric space $\bar{s}$ (middle), a forward model ($\delta$) is learned in the latent space  $\hat{s}$ (right). (b) High-level and low-level planners. The high-level planner generates coarse goals ($\hat{s}_y$) to be used as targets for low-level continuous planner. The dashed line indicates the expected trajectory after $\hat{s}_y$ is reached.}
    \label{fig:architecture}
\end{figure*}

In this paper, we present two contributions. \emph{First}, we develop latent representations that accurately abstract state reachability in the quest towards sample-efficient planning from rich observations. We call this new approach \emph{Plannable Continuous Latent States} or \emph{\methodname} which is a map from observations to latent representation and 
associates neighboring states together by optimizing a contrastive objective inspired by the likelihood function of a Gaussian random walk. The Gaussian is a reasonable model for random exploration \emph{in the embedding space}. Figure~\ref{fig:architecture} shows an overview of our approach, with a specific choice of the initial latent representation based on inverse dynamics.


We hypothesize that \methodname representations are better aligned with the reachability structure of the environment. Our experiments validate that these representations improve the performance of reward-based policy learning and reward-free task completion schemes.
One key benefit of this representation is that it can be used to construct a discretized model of the environment and enable model-based planning to reach an arbitrary state from another arbitrary state. A discretized model (in combination with a simple local continuous planner) can also be used to solve more complex planning tasks that may require combinatorial solvers, like planning a tour across several states in the environment. Similarly to other latent state learning approaches, the learned representations can be used to drive more effective exploration of new states \citep{machado2017eigenoption,hazan2019provably,jinnai2019exploration,amin2021survey}. 

\emph{Secondly}, since the distance in the \methodname representation corresponds to the number of transitions between states, discretizing states at different levels of granularity gives rise to different levels of state abstraction. We hypothesize these abstractions can be efficiently used for hierarchical planning. This is validated in our experiments \footnote{Code for reproducing our experimental results can be found at \href{https://github.com/shivakanthsujit/pclast}{\color{blue}{https://github.com/shivakanthsujit/pclast}}} where we show using multiple levels of hierarchy leads to substantial speed-ups in plan computation.


\section{Related Work}
\label{sec:related_work}

Our work relates to challenges in representation learning for forward/inverse latent-dynamics and using it for ad-hoc goal conditioned planning. We next discuss each of these aspects.\looseness=-1

\textbf{Representation Learning.} Learning representations can be decomposed into \emph{reward-based} and \emph{reward-free} approaches. The former involves both model-free and model-based methods. Model-free methods \citep[e.g.,][]{mnih2013playing} directly learn a policy with rich observation as input. One can consider the penultimate layer as a latent-state representation. Model-based approaches \citep[e.g.][]{hafner2019dream} learn policy, value, and/or reward functions along with the representation. These end-to-end approaches induce task-bias in the representation which makes them unsuitable for diverse tasks. In \emph{reward-free} approaches, the representation is learned in isolation from the task. This includes model-based approaches \citep{ha2018world}, which learn a low-dimensional auto-encoded latent-representation. To robustify, contrastive methods ~\citep{laskin2020curl} learn representations that are similar across positive example pairs, while being different across negative example pairs. They still retain exogenous noise requiring greater sample and representational complexity. This noise can be removed
from latent state by methods like ACRO \citep{islam2022agent} which learns inverse dynamics \citep{mhammedi2023representation}. These \emph{reward-free} representations tend to generalize better for various tasks in the environment. The prime focus of discussed reward-based/free approaches is learning a representation robust to observational/distractor noise; whereas not much attention is paid to enforce the geometry of the state-space. Existing approaches hope that such geometry would emerge as a result of end-to-end training. We hypothesize lack of this geometry affects sample efficiency of learning methods. Temporal contrastive methods (such as \textsc{HOMER}~\cite{misra2020kinematic} and \textsc{DRIML}~\cite{mazoure2020deep}) attempt to address this by learning representations that discriminate among adjacent observations during rollouts, and pairs random observations~\cite{wang2015unsupervised,nair2022r3m}. However, this is still not invariant to exogenous information~\citep{efroni2021provably}.

\textbf{Planning.} Gradient descent methods abound for planning in learned latent states. For example, UPN \citep{srinivas2018universal} applies gradient descent for planning. For continuous latent states and actions, the cross-entropy method (CEM) \citep{rubinstein1999cross}, has been widely used as a trajectory optimizer in model-based RL and robotics \citep{finn2017deep,wang2019exploring,hafner2019learning}. Variants of CEM have been proposed to improve sample efficiency by adapting the sampling distribution of ~\citet{pinneri2021sample} and integrating gradient descent methods \citep{bharadhwaj2020model}. Here, trajectory optimizers are recursively called in an online setting  using an updated observation. This conforms with model predictive control (MPC) \citep{mattingley2011receding}. However, it s limited by planning horizon and returns suboptimal plans in complex tasks. In our work, we ease planning by generating multi-level representation hierarchy and adopting multi-level planner that uses Dijkstra's graph-search algorithm \citep{Dijkstra1959} for coarse planning in each hierarchy level for subgoal generation. A low-level planner (such as CEM) along with a learned latent world model is used to search action sequences to reach next subgoal.\looseness=-1

\textbf{Goal Conditioned Reinforcement Learning (GCRL).} In GCRL, the goal is specified along with the current state and the objective is to reach the goal in least number of steps. Several efforts have been made to learn GCRL policies \citep{kaelbling1993learning, andrychowicz2017hindsight,nasiriany2019planning, fang2018dher, nair2018visual}. Further, reward-free goal-conditioned latent-state planning requires estimating the distance between the current and goal latent state, generally using Euclidean norm ($\ell_2$) for the same. However, it is not clear whether the learned representation is suitable for $\ell_2$ norm and may lead to infeasible/non-optimal plans; even if one has access to true state. So, either one learns a new distance metric \citep{tian2020model, mezghani2023learning, wang2023optimal} which is suitable for the learned representation or learns a representation suitable for the $\ell_2$ norm. In our work, we focus on the latter. Further, GCRL reactive policies often suffer over long-horizon problems which is why we use hierarchical planning on learned latent state abstractions as discussed earlier.
\section{\methodname: Discovery, Representation, and Planning}
\label{sec:method}
In this section, we discuss learning the \methodname representation, constructing a transition model, and implementing a hierarchical planning scheme.

\subsection{Notations and Preliminaries.}

In our work, we extend Exogenous Block Markov Decision Process (EX-BMDP) ~\citep{efroni2021provably} with continuous state and action spaces. We begin by introducing BMDP ~\citep{du2019provably} and then discuss its extension with exogenous (EX) noise. In our discussion, indices of time like $t, t_0, \tau$ will always be integers and $\tau \gg t > t_0$. The Euclidean norm of a matrix $X$ is denoted as $\|X\|$.

\textbf{BMDP.} A BMDP is a tuple $(\mathcal{X},\mathcal{Z},\mathcal{A},T,q,R,\mu)$. Here, $\mathcal{X}, \mathcal{Z}, \text{ and } \mathcal{A}$ are the spaces of observations, latent states, and actions, respectively. The transition function $(T)$ is defined over the latent states as $T \colon \mathcal{Z} \times \mathcal{A} \rightarrow \mathcal{Z}$. Observations are sampled using the emission function $q \colon \mathcal{Z} \rightarrow \mathcal{X}$, with initial latent state distribution given by $z_0 \sim \mu(\cdot)$. The reward is given as $R \colon \mathcal{X} \times \mathcal{A} \rightarrow \mathbb{R}$. In contrast to MDPs, BMDP requires \emph{block assumption} ~\citep{du2019provably}, i.e., emission distribution of any two latent states is disjoint.

\textbf{EX-BMDP.} An EX-BMDP is a BMDP whose latent states can be decoupled into two parts $z=(s,\xi)$, where $s \in \mathcal{S}$ is an endogenous state and $\xi \in \Xi$ is the exogenous state. Further, the transition function and initial distribution can be decoupled as $T(z'|z,a) = T(s'|s,a)T_{\xi}(\xi'|\xi)$ and $\mu(z) = \mu(s)\mu_{\xi}(\xi)$, respectively.

\subsection{ACRO: Learning Endogenous State}
We learn endogenous state representation ($\hat{s}$) using an encoder $\phi$ and an action-controllable (AC) multi-step inverse dynamics $f_{\mathrm{AC}}$ as done in ACRO \citep{islam2022agent}. The encoder $\phi:\mathcal{X}\to\mathcal{S}$ maps high-dimensional images to low-dimensional representation and the inverse dynamics model $f_{AC}: \mathcal{S} \times \mathcal{S} \times [K_{max}] \rightarrow \mathcal{A}\space$
predicts the likelihood of the next action ($a_t$) between a pair of states separated by $k$ steps, i.e., $\PP(a_t | \phi(x_t), \phi(x_{t+k}))$ (we assume that this conditional distribution is Gaussian with a fixed variance).
The functions $\phi$ and $f_{AC}$ are optimized together using \cref{eq:multi_inverse_obj_loss,eq:multi_inverse_obj} below, where $t \sim U(1,\mathcal{T})$ is the index of time, and $k \sim U\left(1,K_{max}\right)$ is the amount of look-ahead steps with $K_{max}$ as the diameter of the control-endogenous MDP~\citep{lamb2022guaranteed,islam2022agent}:
\begin{subequations}%
\begin{align}
\notag
&
    \mathcal{L}_{s}(\phi, f_{\mathrm{AC}}, x_t, a_t, x_{t+k}, k)
\\&\qquad{}
        = \|a_t - f_{\mathrm{AC}}(\phi(x_t), \phi(x_{t+k}); k)\|^2,
 \label{eq:multi_inverse_obj_loss}
\\[2pt]
&
\arg\min_{\phi} \min_{f_{\mathrm{AC}} }\mathop{\mathbb{E}}_{\substack{t}} \mathop{\mathbb{E}}_{k} \mathcal{L}_{s} \left(\phi,f_{\mathrm{AC}},x_t,a_t,x_{t+k},k\right)
\label{eq:multi_inverse_obj}
\end{align}
\end{subequations}

\subsection{Learning the \methodname map} 
\label{sec:map_learning}
While the encoder $\phi$ and the inverse dynamics model $f_{AC}$ are designed to filter out the exogenous noise, they do not lead to representations that reflect the reachability structure (see \cref{fig:overview-acro-clustering}). To enforce states' reachability, we learn a map $\psi: S \rightarrow \bar{S}$, which associates nearby states ($s \in S$) to have similar representation in $\bar{s} \in \bar{S}$, based on transition deviations. 

Learning $\psi$ is inspired by local random exploration that enforces a Gaussian random walk in the embedding space ($\bar{S}$). This allows states visited in fewer transitions to be closer to each other. A Gaussian random walk with variance $\sigma I$ (where $I$ is an identity matrix) for $k$ steps in $\bar{S}$ would induce a conditional distribution $\mathbb{P}(\bar{s}_{t+k}| \bar{s_t})\propto\exp{\bigl\{-\frac{\|\bar{s}_{t+k}-\bar{s}_t\|^2}{2k\sigma^2}\bigr\}}$. \emph{This also requires a reversibility assumption that there exists a k-step path from $s_{t+k}$ to $s_t$.} Instead of fitting $\psi$ to this likelihood directly, we fit a contrastive version, based on the following process for generating triples $\langle y,\bar{s}_t,\bar{s}_{t+k}\rangle$. First, we flip a random coin with outcome $y\in\{0,1\}$ and then predict $y$ using $\bar{s}_t$ and $\bar{s}_{t+k}$, yielding the likelihood function
\begin{align}
    \PP_k(y = 1 | \bar{s}_t, \bar{s}_{t+k}) = \sigma(e^\alpha - e^\beta \|\bar{s}_t - \bar{s}_{t+k}\|),
\end{align} 
for suitable $\alpha$ and $\beta$ (see the derivation in Appendix~\ref{app:theory}). We use $e^\alpha$ and $e^\beta$ to smoothly enforce positive values.


We employ a contrastive learning loss $\mathcal{L}_\psi$ in \cref{eq:lns_1,eq:lns_2,eq:lns_3,eq:lns_4} to fit $\psi$ as well as the parameters $\alpha$ and $\beta$ by averaging over the expected loss. In $\mathcal{L}_\psi$, $t \sim U(1, \mathcal{T})$, $r \sim U(1,\mathcal{T})$,  $d \sim U\left(1,d_m\right)$ for a hyperparameter $d_m$. Positive examples ($x_t$ and $x_{t+d}$) are drawn for the contrastive objective uniformly over $d_m$ steps and optimized using \cref{eq:lns_1}. Negative examples ($x_r$) are sampled uniformly from a data buffer and optimized with \cref{eq:lns_2}, which encourages change in state representation to be higher for negative samples.%
\begin{subequations}%
\begin{align}
\notag
&
    \mathcal{L}_{m_+}(\psi, \hat{s}_A, \hat{s}_B, \alpha, \beta)
\\&\qquad{}
         = -\log(\sigma(e^\alpha - e^\beta \|\psi(\hat{s}_A) - \psi(\hat{s}_B)\|^2))
    \label{eq:lns_1}
\\
\notag
&
	\mathcal{L}_{m_-}(\psi, \hat{s}_A, \hat{s}_B, \alpha, \beta)
\\&\qquad{}
    = -\log(1 - \sigma(e^\alpha - e^\beta \|\psi(\hat{s}_A) - \psi(\hat{s}_B)\|^2))
    \label{eq:lns_2}
\end{align}
\begin{align}
\notag
&   \mathcal{L}_\psi(\psi, \phi, \alpha, \beta, x_t, x_{t+d}, x_r)
\\
\notag
&\qquad{}
    = \mathcal{L}_{m_+}(\psi, \phi(x_t), \phi(x_{t+d}), \alpha, \beta)
\\&\qquad\qquad{}
    + \mathcal{L}_{m_-}(\psi, \phi(x_t), \phi(x_r), \alpha, \beta)
    \label{eq:lns_3}
\\[2pt]
&
\arg\min_{\substack{\psi \in \Psi,\\\alpha,\beta \in \mathbb{R}}} \mathop{\mathbb{E}}_{\substack{t,r}} \mathop{\mathbb{E}}_{d}  \mathcal{L}_\psi(\psi, \phi, \alpha, \beta, x_t, x_{t+d}, x_r)
    \label{eq:lns_4}
\end{align}
\end{subequations}

In our approach, $\phi$ is not optimized with respect to the contrastive loss $\mathcal{L}_\psi$.  This is motivated by the work of \citet{efroni2022ppe}, who proved that temporal contrastive objective loss can be reduced by capturing exogenous noise. Hence, optimizing $\phi$ with contrastive loss may lead it to acquire information on the exogenous state.  To avoid this failure case, we task $\phi$ with capturing the agent-centric state and task $\psi$ with learning the local neighborhood structure.

\subsection{Learning a latent forward model} 
\label{sec:learn_for_planning}
In order to plan in our learned latent space, we learn a one-step latent dynamics $\delta : \mathcal{S} \times \mathcal{A} \rightarrow \mathcal{S}$. This estimates observational dynamics by predicting $\phi(x_{t+1}) \approx \delta(\phi(x_t), a_t)$. The forward model $\delta$ is parameterized as a fully-connected network of a parameterized family $\mathcal{F}$, optimized with the following objective:
\begin{subequations}%
	\begin{align}
		&\mathcal{L}_\delta(\delta, x_t, a_t, x_{t+1}) = \|\phi(x_{t+1}) - \delta(\phi(x_t), a_t)\|^2,\\
		&\arg\min_{\delta \in \mathcal{F}} \mathop{\mathbb{E}}_t \mathcal{L}_\delta(\delta, x_t, a_t, x_{t+1})
	\end{align}
\end{subequations}

In our approach, we jointly optimize all of our architecture components $\phi(\cdot)$, $f_{AC}(\cdot)$, $\psi(\cdot)$, and $\delta(\cdot,\cdot)$.

\subsection{Planning}
We describe utility of \methodname for abstraction and generating goal-conditioned abstract plans.

\textbf{High-Level Planner.} 
Let $\hat{s_t} = \phi(x_t)$ denote the latent state. In the planning problem, we aim to navigate the agent from an initial latent state $s_{init}$ to a target latent state $s_{goal}$ following the latent forward dynamics $\hat{s}_{t+1} = \delta(\hat{s}_t, a_t)$. Since $\delta$ is highly nonlinear, it presents challenges for use in global planning tasks. Therefore, we posit that a hierarchical planning scheme with abstractions can improve the performance and efficacy of planning by providing waypoints for the agent to track using global information of the environment. 

To find a waypoint $\hat{s}^*$ in the latent space, we first divide the latent space into $C$ clusters by applying $k$-means to an offline dataset $D_{\psi}$, which is created by $\methodname$ transformation of offline observations, i.e., $D_\psi = \{\dotsc,\psi(\phi(x)),\dotsc\}$. We use $\{c_i\}_{i=1}^C$ to denote each cluster. Our assumption of reversibility described in \cref{sec:map_learning} ensures a path exists between any two states of the same cluster.
For each cluster, we store the latent-state representation of its centroid alongside the \methodname transformation of this latent state.


An abstraction of the environment is given by a graph $\mathcal{G}$ with nodes $\{c_i\}_{i=1}^C$ and edges defined by the reachability of each cluster, i.e., an edge from node $c_i$ to node $c_j$ is added to the graph if there are transitions of latent states from cluster $c_i$ to cluster $c_j$ in the offline transition dataset. On the graph $\mathcal{G}$, we apply Dijkstra's shortest path algorithm~\citep{Dijkstra1959} to find the next cluster the agent should go to and choose the latent state corresponding to centroid of that cluster as the waypoint $\hat{s}^*$. This waypoint is passed to a low-level planner to compute the action.  

\textbf{Low-Level Planner.}
Given the current latent state $\hat{s}_0$ and the waypoint $\hat{s}^*$ to track, the low-level planner finds the action to take by solving a trajectory optimization problem using the cross-entropy method (CEM)~\citep{de2005tutorial}. The details are shown in Appendix~\ref{subsec:appendix-low-level-planner}.

\textbf{$n$-Level Planner.}
To improve the efficiency of finding the waypoint $\hat{s}^*$, we propose to build a hierarchical abstraction of the environment such that the high-level planner can be applied at different levels of granularity, leading to an overall search time reduction of Dijkstra's shortest path algorithm. The \emph{$n$-Level Planner} creates $n$ abstraction levels, indexed by $i=1,\dotsc,n$, from finest to coarsest, with $i=1$ corresponding to low-level planner.\footnote{When $n=1$, we only apply the low-level planner without searching for any waypoint.} At level $i\ge 2$, we partition the latent space into $C_i$ clusters using $k$-means, and we have $C_2 > C_3 > \dotsb > C_n$. For each abstraction level, we construct the discrete transition graph $\mathcal{G}_i$ accordingly, which is used to search for the waypoint $\hat{s}^*$ with increasing granularity as shown in Algorithm~\ref{alg:mpc}. This procedure guarantees that the start and end nodes are always a small number of hops away in each call of Dijkstra's algorithm. In Section~\ref{subsec:impact-multi-level-abstraction-planning}, our experiments show that multi-level planning leads to a significant speedup compared with using only the finest granularity.


\begin{algorithm}[tbh!]
\caption{$n$-Level Planner}
\label{alg:mpc}
\begin{algorithmic}[1]

\REQUIRE~~\\
Current observation $x_t$\\
Goal observation $x_{goal}$\\
Planning horizon $H$\\
Encoder $\phi(\cdot)$\\
\methodname map $\psi(\cdot)$\\
Latent forward dynamics $\delta(\cdot, \cdot)$\\
Multi-Level discrete transition graphs $\{\mathcal{G}_i\}_{i=2}^n$
\ENSURE Action sequence $\{a_i\}_{i=0}^{H-1}$
\STATE Compute current continuous latent state $\hat{s_t} = \phi(x_t)$ and target latent state $\hat{s}^* = \phi(x_{goal})$. \\
    \COMMENT{See Appendix~\ref{app:level} for details of high-level planner and low-level planner.}
\FOR{$i=n, n-1, \dotsc, 2$}
\STATE $\hat{s}^*$ = high-level planner($\hat{s}_t$, $\hat{s}^*$, $\mathcal{G}_i$) \\
\COMMENT{Update waypoint using a hierarchy of abstraction.}
\ENDFOR
\STATE $\{a_i\}_{i=0}^{H-1} =$ low-level planner($\hat{s}_t, \hat{s}^*, H, \delta, \psi$) \\
\COMMENT{Solve the trajectory optimization problem.}
\end{algorithmic}
\end{algorithm}

\section{Experiments}
\label{sec:experiments}
In this section, we address the following questions via experimentation over environments of different complexities: (1)~Does the \methodname representation lead to performance gains in reward-based and reward-free goal-conditioned tasks? (2)~Does increasing abstraction levels lead to more computationally efficient and better plans? (3)~What is the effect of \methodname map on abstraction?

\subsection{Environments}
\label{subsec:environments}
\begin{figure}[h!]
    \centering
    \begin{subfigure}[t]{0.15\textwidth}
         \centering
         \includegraphics[width=\textwidth, height=0.68\textwidth]{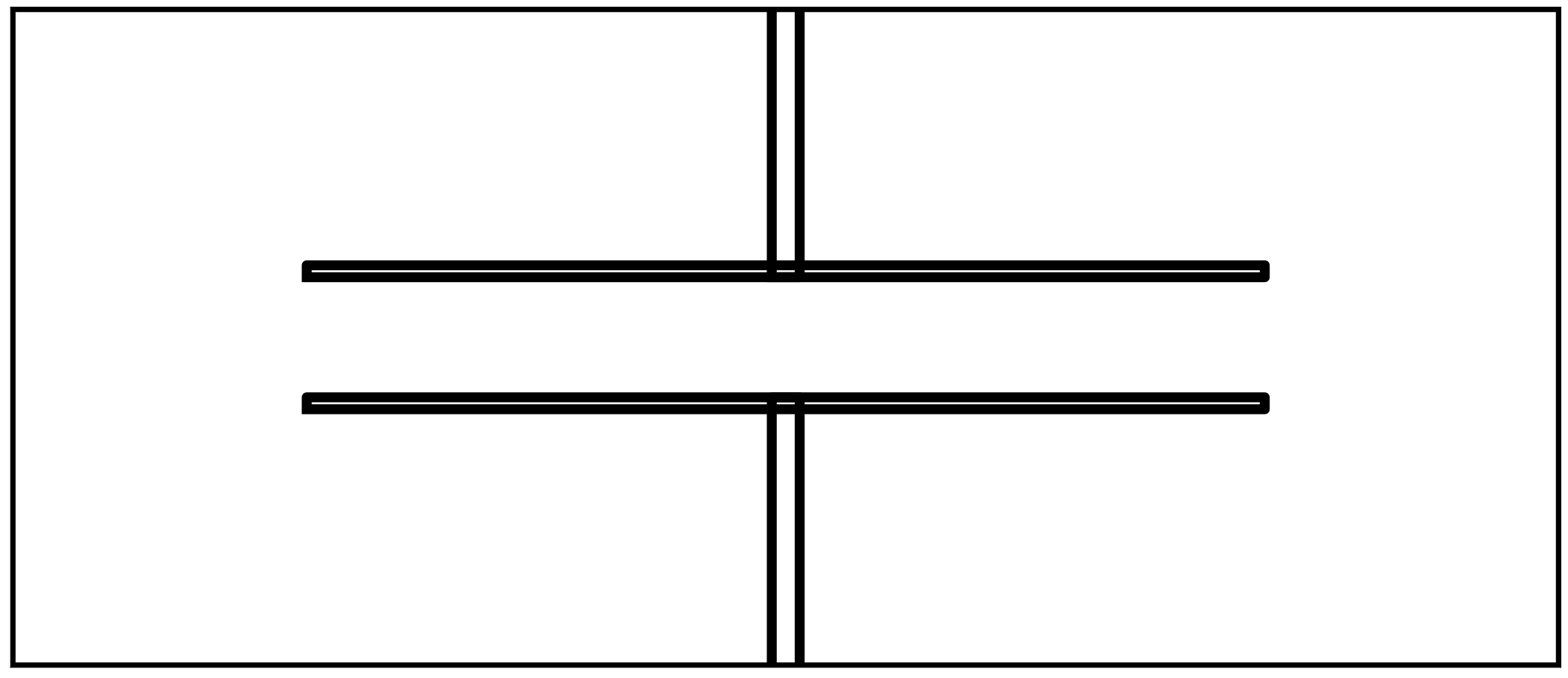}
         \caption{Hallway}
         \label{fig:Hallway}
    \end{subfigure}
    \begin{subfigure}[t]{0.15\textwidth}
         \centering
         \includegraphics[width=\textwidth, height=0.68\textwidth]{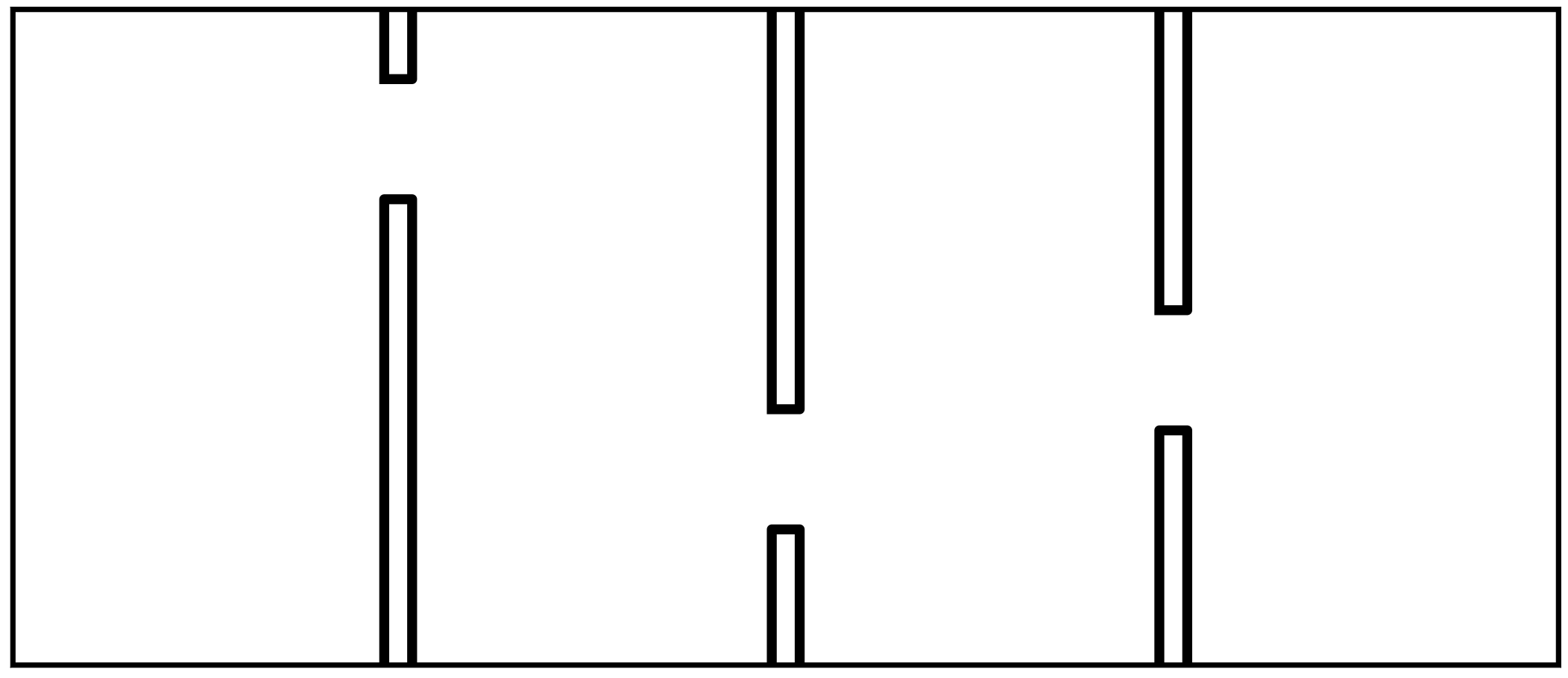}
         \caption{Rooms}
         \label{fig:Rooms}
    \end{subfigure}
    \begin{subfigure}[t]{0.15\textwidth}
         \centering
         \includegraphics[width=\textwidth, height=0.68\textwidth]{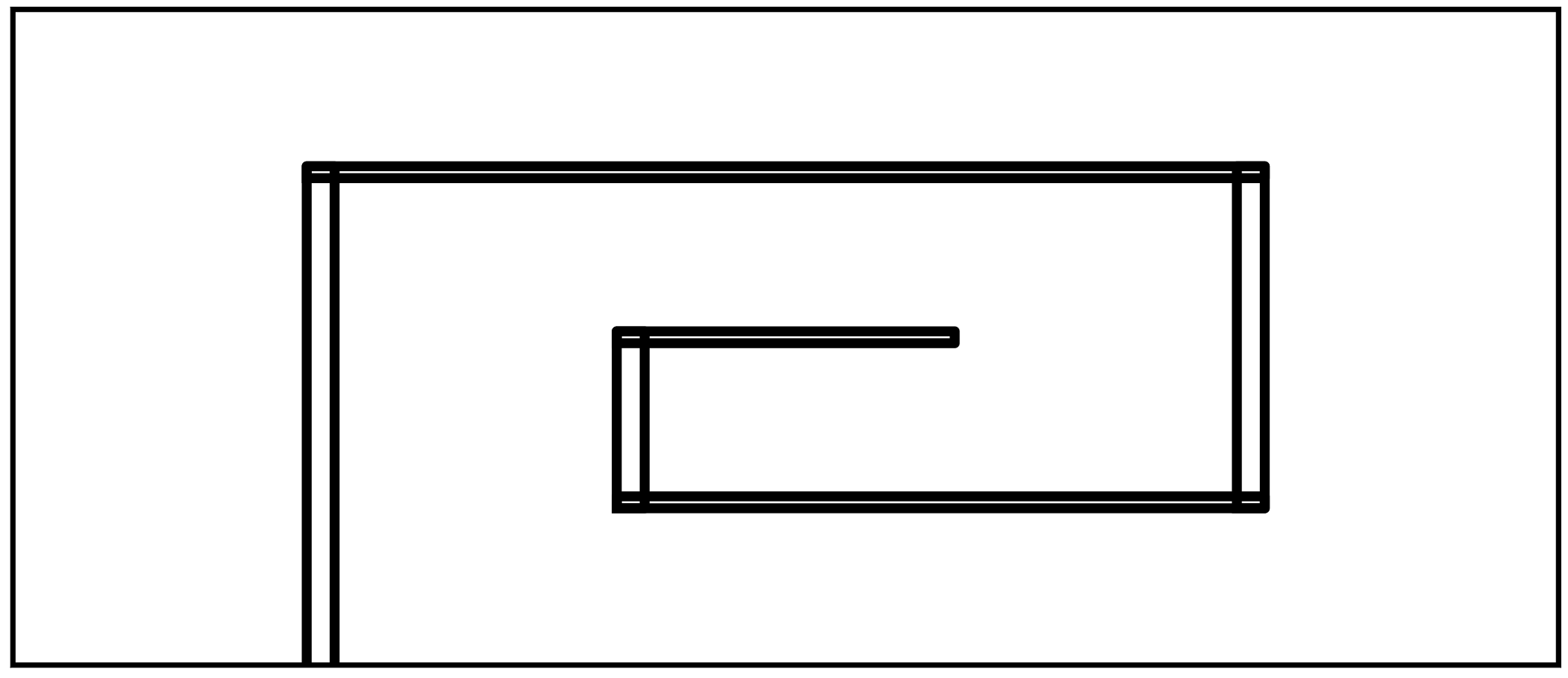}
         \caption{Spiral}
         \label{fig:Spiral}
    \end{subfigure}
    \begin{subfigure}[t]{0.25\textwidth}
         \centering
         \includegraphics[width=\textwidth]{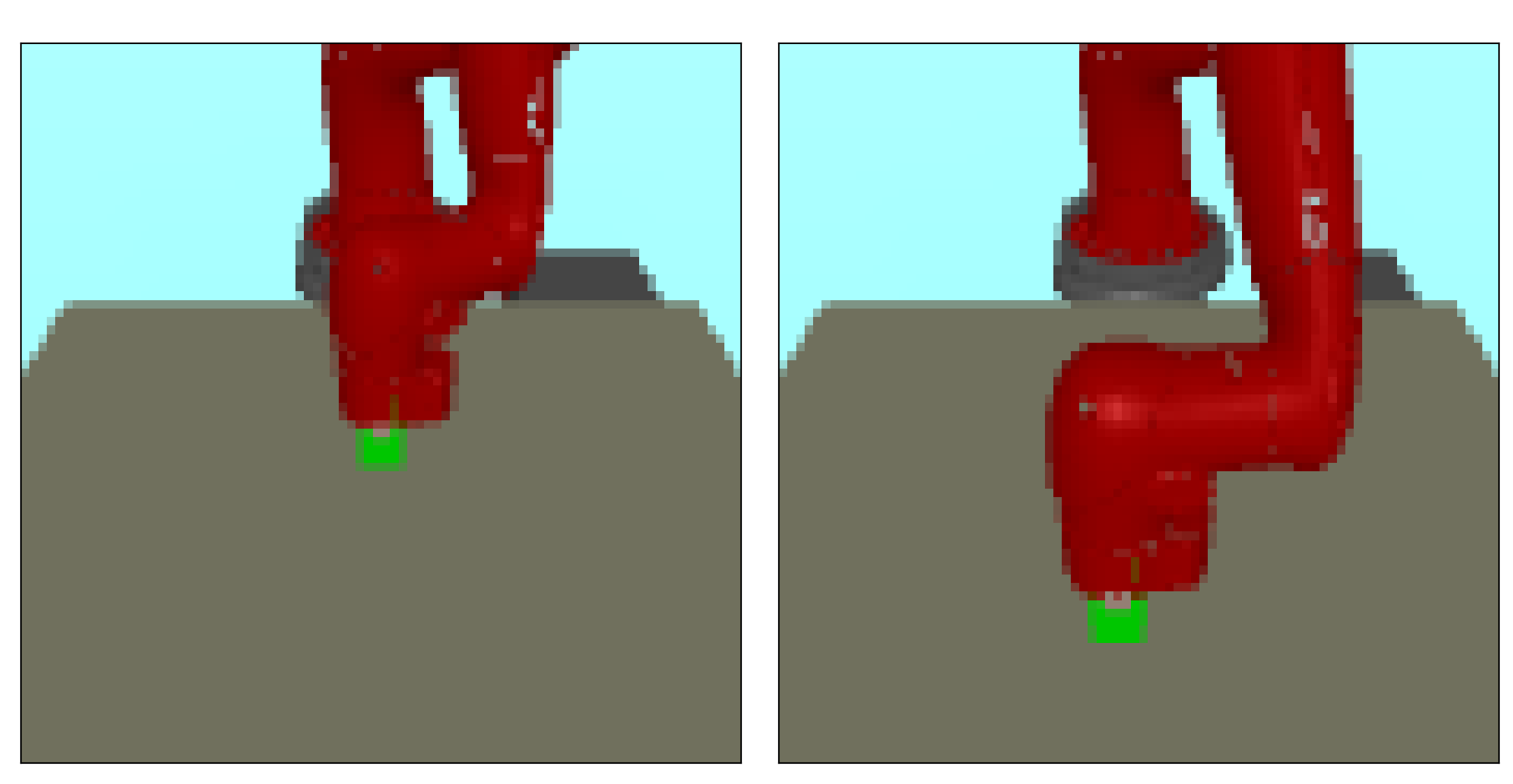}
         \caption{Sawyer Reach Environment}
         \label{fig:robot_arm}
    \end{subfigure}
    \caption{Environments: (a), (b) and (c) show different wall configurations of \emph{Maze2d} environment for point-mass navigation task and (d) shows top-down view of robot-arm environment with the task of reaching various goal positions in 2D-planar space. 
    }
    \label{fig:maze_environment}
\end{figure}

We consider three categories of environments for our experiments and discuss them as follows:

\textbf{Maze2D---Point Mass.} We created 2D maze point-mass environments with continuous actions and states. The environments comprise of different wall configurations with the goal of navigating a point mass. The size of the grid is $(100 \times 100)$ and each observation is a 1-channel image of the grid with ``0" marking an empty location and ``1" marking the ball's coordinate location $(x,y)$. Actions comprise of $(\Delta x, \Delta y)$ and specify the coordinate space change by which the ball should be moved. This action change is bounded by $[-0.2,0.2]$. There are three different maze variations: \textsc{Maze-Hallway}, \textsc{Maze-Spiral}, and \textsc{Maze-Rooms} whose layouts are shown in \cref{fig:maze_environment}(a, b and c). Further, we have dense and sparse reward variants for each environment, details of which are given in \cref{app:2dmaze}. We created an offline dataset of 500K transitions using a random policy for each environment which gives significant coverage of the environment's state-action space.

\textbf{Robotic Arm.} We extended our experiments to the \emph{Sawyer-Reach} environment of \citet{nair2018visual} (shown in Fig. \ref{fig:robot_arm}). It consists of a 7 DOF robotic arm on a table with an end-effector. The end-effector is constrained to move only along the planar surface of the table. The observation is an $84$-by-$84$ RGB image of the top-down view of the robotic arm and actions are 2-dimensional continuous vectors that control the end-effector coordinate position. The agent is tested on its ability to control the end-effector to reach random goal positions. The goals are given as images of the robot arm in the goal state. Similar to maze2d environment, we generate an offline dataset of 20K transitions using rollouts from a random policy. Likewise to maze, it has dense and sparse reward variants.\looseness=-1

\textbf{Exogenous Noise Mujoco.} We adopted control-tasks \emph{``Cheetah-Run"} and \emph{``Walker-walk"} from visual-d4rl \citep{vd4rl} benchmark which provides offline transition datasets of various qualities. The datasets include high-dimensional agent tracking camera images, to which exogenous noise is added by concatenating randomly sampled images from another distribution as shown in \cref{fig:offlineRL_observations_main} and discussed further in \cref{appendix:exo_offline}. We consider \emph{``medium, medium-expert, and expert"} datasets and use \emph{``random"} dataset of same domain as source of exogenous noise. The general objective in these tasks is to keep the agent alive and move forward based on images with exogenous noise.

\begin{figure}[tbh!]
    \centering
    \includegraphics[scale=0.4]{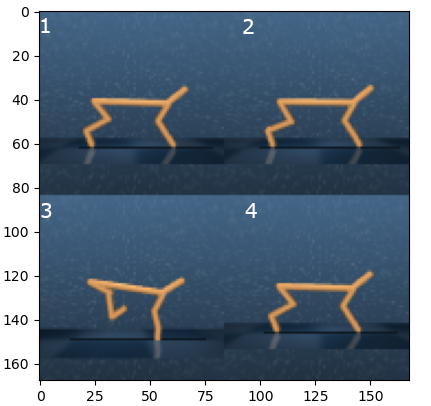}
    \caption{Illustration of an observation for \emph{Cheetah-Run}, where the controllable environment image (1) is placed along with exogenous noise images (2-4) in a $4 \times 4$ grid. Numbers on images are for reference only. This grid of images is given as input to agent.
    }
    \label{fig:offlineRL_observations_main}
\end{figure}

\subsection{Impact of representation learning on goal-conditioned RL}
\label{subsec:representation-goal-rl}
We investigate the impact of different representations on performance in goal-conditioned model-free methods. First, we consider methods which use explicit-reward signal for representation learning. As part of this, we trained goal-conditioned variant of PPO \citep{schulman2017proximal} on each environment with different current state and goal representation methods. This includes: (1) Image representation for end-to-end learning, (2) ACRO representation~\cite{islam2022agent}, and (3) \methodname representation. For (1), we trained PPO for 1 million environment steps. For (2) and (3), we first trained representation using an offline dataset and then used frozen representation with PPO during online training for 100K environment steps only. In the case of \emph{Sawyer-Reach}, we emphasize the effect of limited data and reserved experiments to 20K online environment steps. We also did similar experiment with offline CQL \citep{kumar2020conservative} method with pre-collected dataset.

Secondly, we consider RL with Imagined Goals (RIG) \citep{nair2018visual}, a method which \emph{doesn't need an explicit reward signal} for representation learning and planning. It is an online algorithm which first collects data with simple exploration policy. Thereafter, it trains an embedding using VAE on images and fine-tunes it over the course of training. Goal-conditioned policy and value functions are trained over the VAE embedding of goal and current state. The reward function is the negative of $\ell_2$ distance in the latent representation of current and goal observation. In our experiments, we consider pre-trained ACRO and \methodname representation in addition to default VAE representation. Pre-training was done over the datasets collected in \cref{subsec:environments}.

Our results in \cref{tab:expr_result} show PPO and CQL have poor performance when using direct images as representations in maze environments. However, ACRO and \methodname representations improve performance. Specifically, in PPO, \methodname leads to significantly greater improvement compared to ACRO for maze environments; training curves for the same are shown in \cref{fig:ppo-comparison} (Appendix). This suggests that enforcing a neighborhood constraint facilitates smoother traversal within the latent space, ultimately enhancing goal-conditioned planning. \methodname in CQL gives significant performance gain for \emph{Maze-Hallway} over ACRO; but they remain within standard error of each other in \emph{Maze-Rooms} and \emph{Maze-Spiral}. Generally, each method does well on \emph{Sawyer-Reach} environment. We assume it is due to lack of obstacles which allows a linear path between any two positions easing representation learning and planning from images itself. In particular, different representations tend to perform slightly better in different methods such as ACRO does better in PPO (sparse), \methodname does in CQL, and image itself does well in PPO (dense) and RIG.

\begin{table*}[ht!]
\caption{Impact of different representations on policy learning and planning. The numbers represent mean and standard error of the percentage success rate of reaching goal states, estimated over 5 random seeds. RIG and $n$-Level Planner do not use an external reward signal. In $n$-Level Planner, we use $n=5$ abstraction levels. Best mean performance in each task is highlighted in bold.\looseness=-1}
\label{tab:expr_result}
\vskip 0.15in
\begin{center}
\begin{small}
\begin{sc}
    \begin{tabular}{|c|c|c|c|c|c|}
        \hline
         Method & Reward type & Hallway & Rooms & Spiral & Sawyer-Reach \\
         \hline
         \hline
          PPO &  Dense & 6.7 $\pm$ 0.6 & 7.5 $\pm$ 7.1 & 11.2 $\pm$ 7.7 & \textbf{86.00 $\pm$ 5.367} \\
         PPO + ACRO &  Dense & 10.0 $\pm$ 4.1 & 23.3 $\pm$ 9.4 &  23.3 $\pm$ 11.8 & 84.00 $\pm$ 6.066\\
         PPO + \methodname &  Dense & \textbf{66.7 $\pm$ 18.9} & \textbf{43.3 $\pm$ 19.3} &  \textbf{61.7 $\pm$ 6.2} & 78.00 $\pm$ 3.347 \\
         \hline
        \hline
        PPO &  Sparse & 1.7 $\pm$ 2.4 & 0.0 $\pm$ 0.0 & 0.0 $\pm$ 0.0 & 68.00 $\pm$ 8.198\\
         PPO + ACRO &  Sparse & 21.7 $\pm$ 8.5 & 5.0 $\pm$ 4.1 & 11.7 $\pm$ 8.5 & \textbf{92.00 $\pm$ 4.382}\\
         PPO + \methodname &  Sparse & \textbf{50.0 $\pm$ 18.7} & \textbf{6.7 $\pm$ 6.2} & \textbf{46.7 $\pm$ 26.2} & 82.00 $\pm$ 5.933 \\
        \hline
        \hline
          CQL &  Sparse & 3.3 $\pm$ 4.7 & 0.0 $\pm$ 0.0 & 0.0 $\pm$ 0.0 & 32.00 $\pm$ 5.93\\
           CQL + ACRO &  Sparse & 15.0 $\pm$ 7.1 & \textbf{33.3 $\pm$ 12.5} & \textbf{21.7 $\pm$ 10.3}  & 68.00 $\pm$ 5.22\\
            CQL + \methodname &  Sparse & \textbf{40.0 $\pm$ 0.5} & 23.3 $\pm$ 12.5 & 20.0 $\pm$ 8.2 & \textbf{74.00 $\pm$ 4.56} \\
         \hline\hline
         RIG & None  & 0.0 $\pm$ 0.0 & 0.0 $\pm$ 0.0 & 3.0 $\pm$ 0.2 & \textbf{100.0 $\pm$ 0.0}\\
           RIG + ACRO & None  & \textbf{15.0 $\pm$ 3.5} & 4.0 $\pm$ 1. & \textbf{12.0 $\pm$ 0.2}  & 100.0 $\pm$ 0.0 \\
          RIG + \methodname & None & 10.0 $\pm$ 0.5 & 4.0 $\pm$ 1.8 & 10.0 $\pm$ 0.1 & 90.0 $\pm$ 5\\
          \hline\hline
           Low-Level Planner + \methodname &  None & 86.7 $\pm$ 3.4 & 69.3$\pm$ 3.4 &  50.0 $\pm$ 4.3 & $\pm$ \\
         $n$-Level Planner + \methodname &  None & \textbf{97.78 $\pm$ 4.91} & \textbf{89.52 $\pm$ 10.21} &  \textbf{89.11 $\pm$ 10.38} & 95.0 $\pm$ 1.54 \\
         \hline
    \end{tabular}
\end{sc}
\end{small}
\end{center}
    \vskip -0.1in
\end{table*}

\subsection{Impact of \methodname on state abstraction}
\label{subsec:impact-state-abstraction}
\begin{figure}[tbh!]
\centering
\begin{minipage}[]{.45\linewidth}
    \begin{subfigure}{\linewidth}
    \centering
    \includegraphics[width=\linewidth]{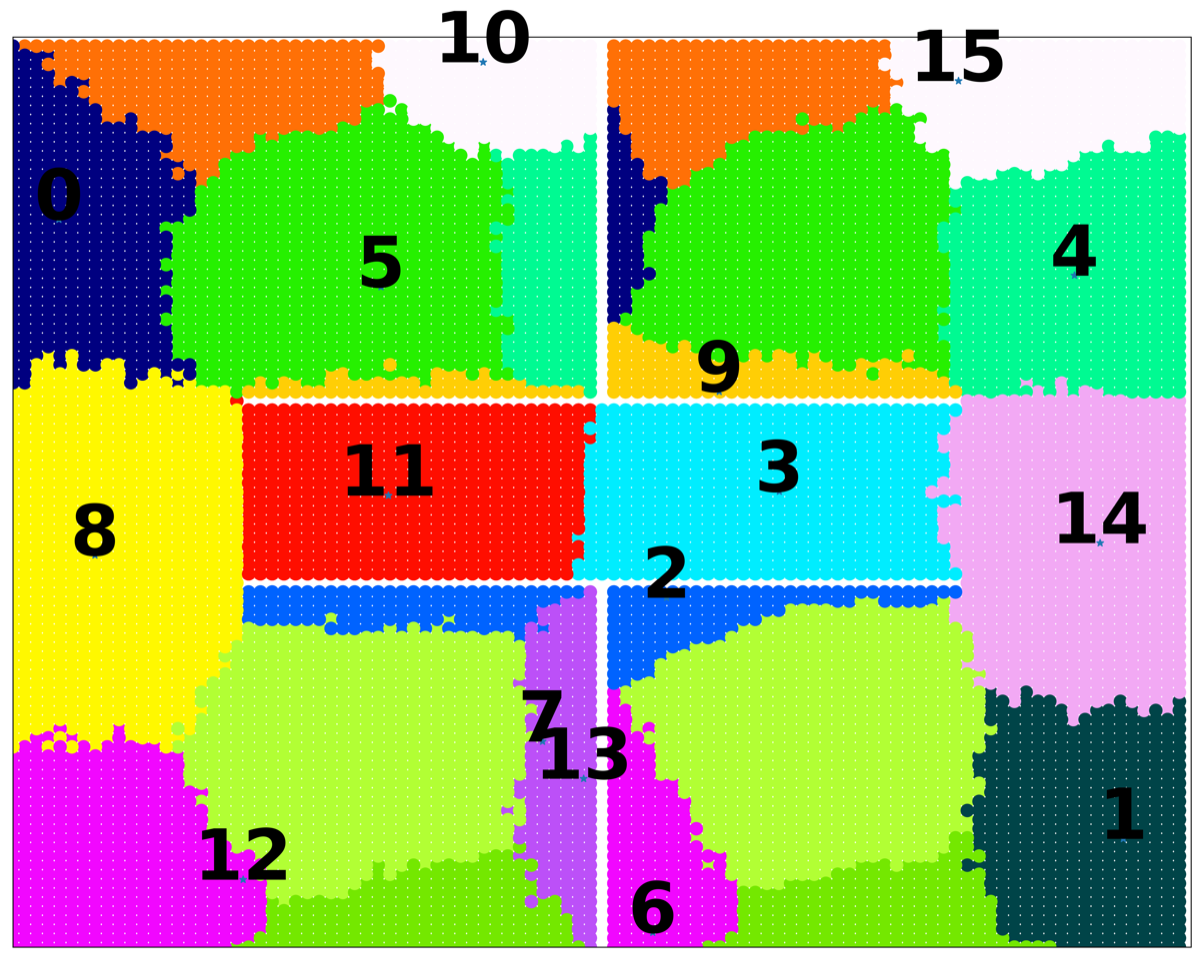}
    \caption{Clusters ACRO}
    \label{fig:maze_experiment_hallway_acro}  
    \end{subfigure}%
\end{minipage}
\begin{minipage}[]{.45\linewidth}
    \begin{subfigure}{\linewidth}
    \centering
    \includegraphics[width=\linewidth]{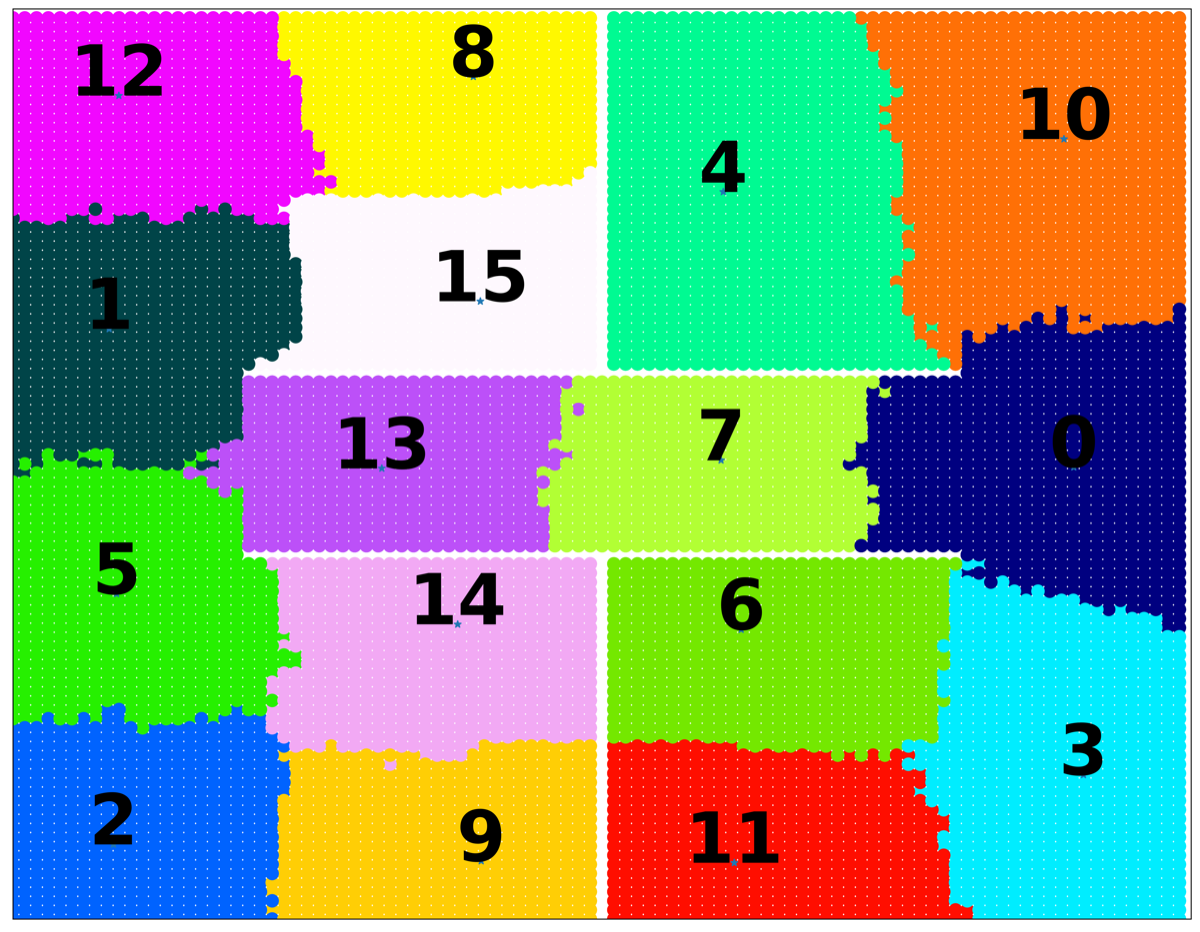}
    \caption{Clusters \methodname}
    \label{fig:maze_experiment_hallway_pclast}  
    \end{subfigure}%
\end{minipage}
\begin{minipage}[]{.45\linewidth}
    \begin{subfigure}{\linewidth}
    \centering
    \includegraphics[width=\linewidth]{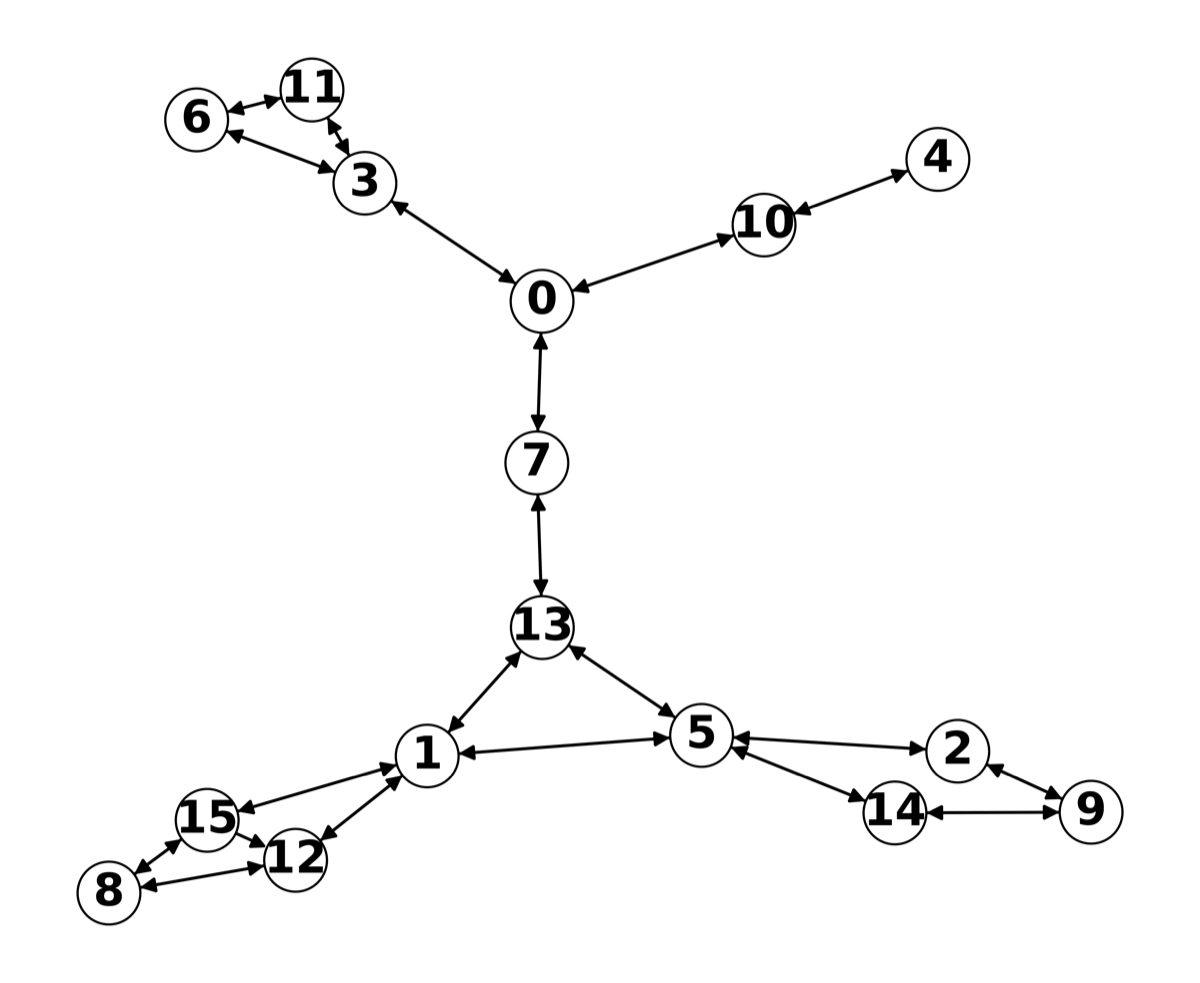}
    \caption{State-transitions \methodname}
    \label{fig:maze_experiment_hallway_state_transitions}  
    \end{subfigure}%
\end{minipage}
\begin{minipage}[]{.45\linewidth}
    \begin{subfigure}{\linewidth}
    \centering
    \includegraphics[width=\linewidth]{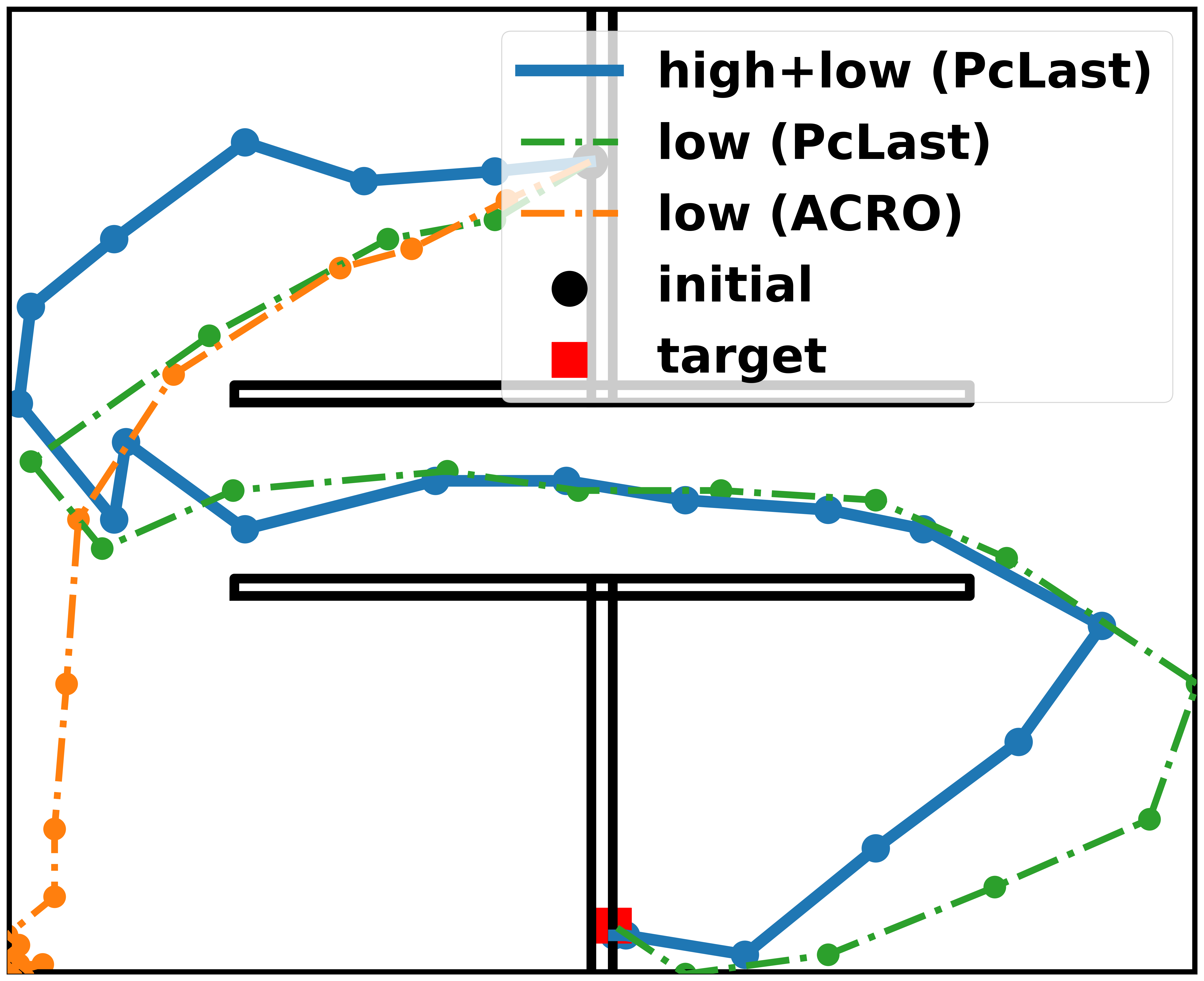}
    \caption{Planning Trajectories}
    \label{fig:maze_experiment_hallway_trajectories_pclast}  
    \end{subfigure}%
\end{minipage}
\caption{
Clustering, Abstract-MDP, and Planning are shown for \emph{Maze-Hallway} environment. In (a) and (b), we show $k$-means ($k=16$) clustering of latent states learned by \methodname and ACRO, respectively. In (c), we show the abstract transition model of the discrete states learned by \methodname (b) which captures the environment's topology. Finally, in (d), we show maze configuration and the executed trajectories of the agent from the initial location (black) to the target location (red) using \emph{n-Level} ($n=2$) planner (blue) with \methodname and just low-level planner with ACRO (orange) and \methodname (green) representation for cost minimization .}
\label{fig:maze_experiment_hallway} 
\end{figure}

\begin{figure}[tbh!]
\centering
\begin{minipage}[]{.45\linewidth}
    \begin{subfigure}{\linewidth}
    \centering
    \includegraphics[width=\linewidth]{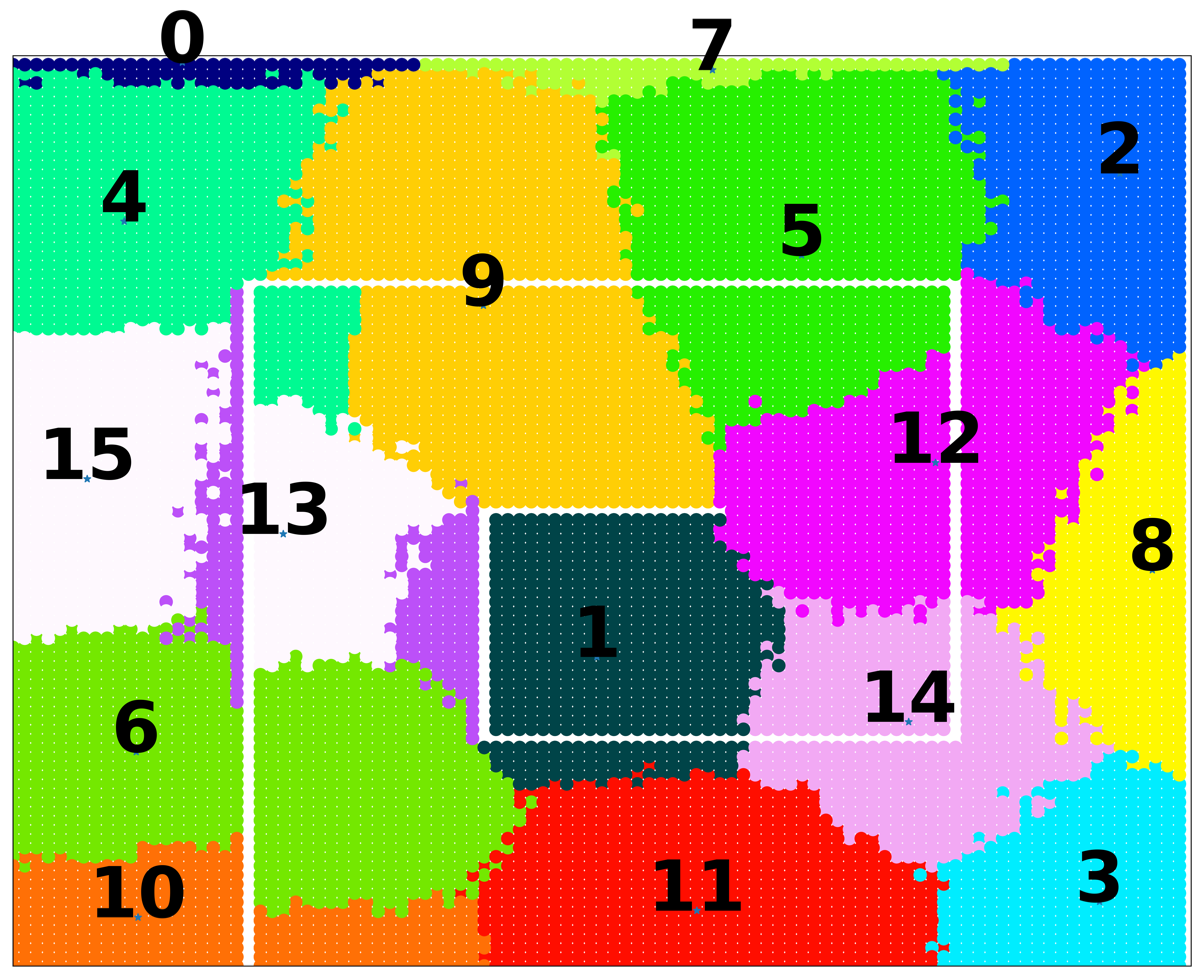}
    \caption{Clusters ACRO}
    \label{fig:maze_experiment_spiral_acro}   
    \end{subfigure}%
\end{minipage}
\begin{minipage}[]{.45\linewidth}
    \begin{subfigure}{\linewidth}
    \centering
    \includegraphics[width=\linewidth]{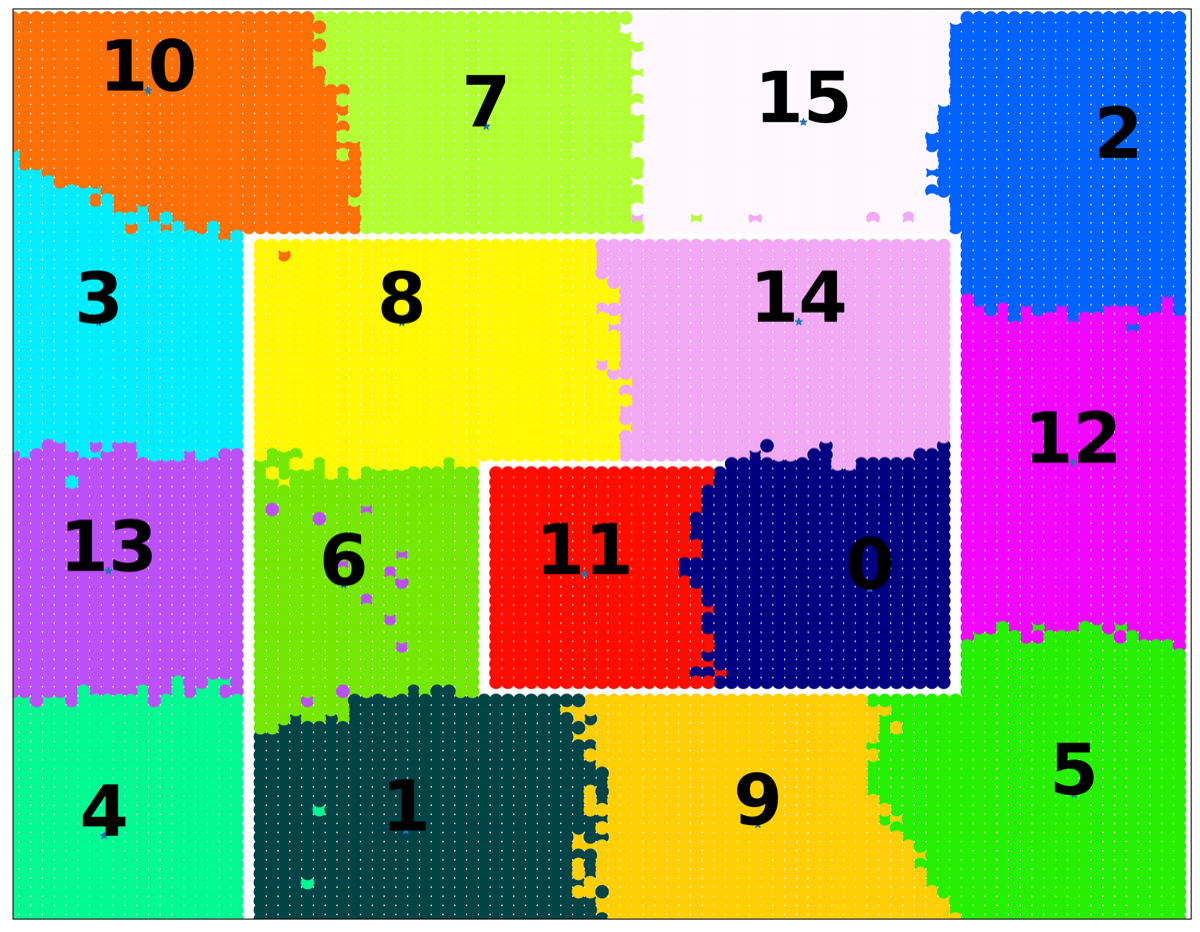}
    \caption{Clusters \methodname}
    \label{fig:maze_experiment_spiral_pclast}   
    \end{subfigure}%
\end{minipage}
\begin{minipage}[]{.48\linewidth}
    \begin{subfigure}{\linewidth}
    \centering
    \includegraphics[width=\linewidth]{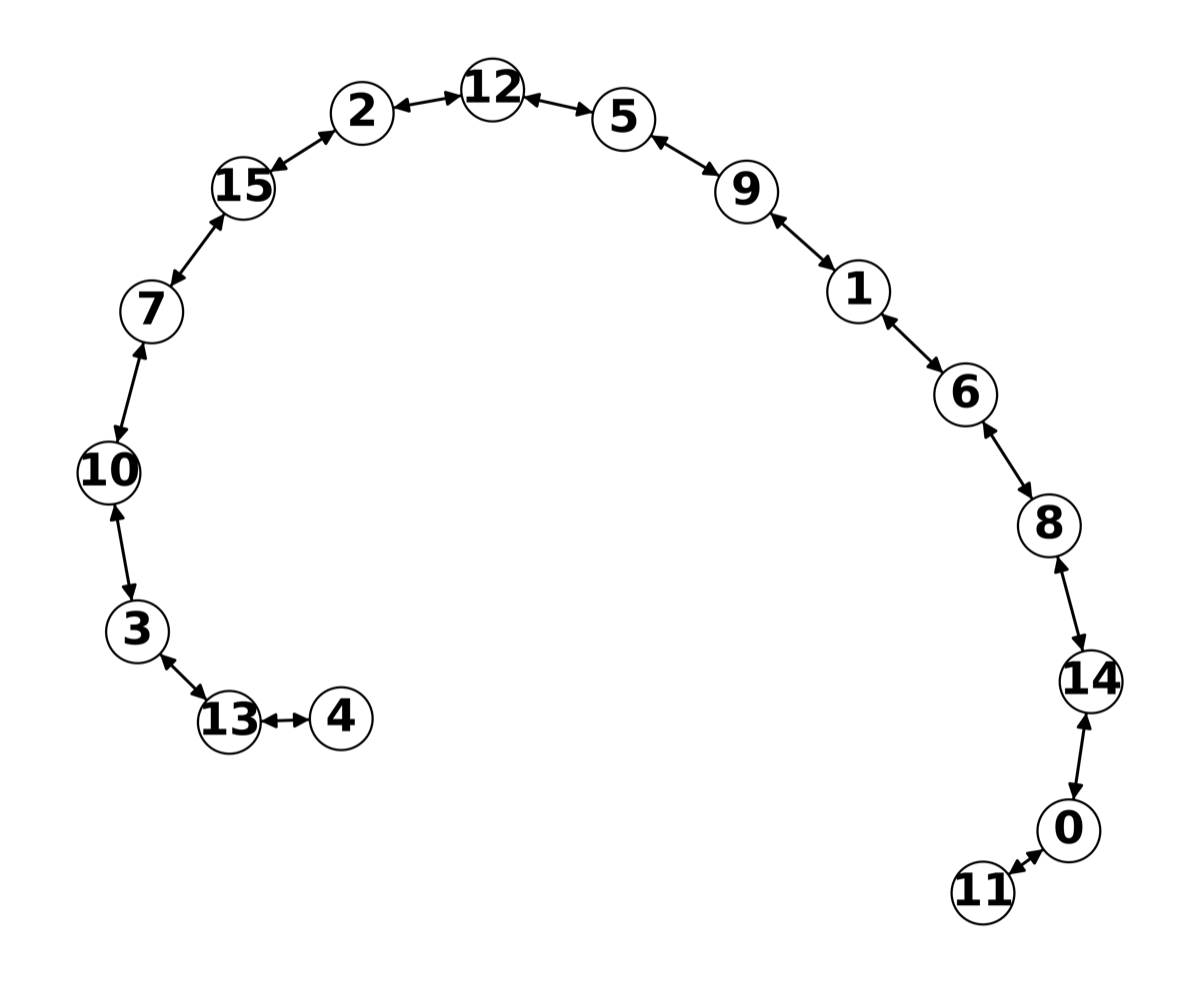}
    \caption{State-transitions \methodname}
    \label{fig:maze_experiment_spiral_state_transitions}  
    \end{subfigure}%
\end{minipage}
\begin{minipage}[]{.45\linewidth}
    \begin{subfigure}{\linewidth}
    \centering
    \includegraphics[width=\linewidth]{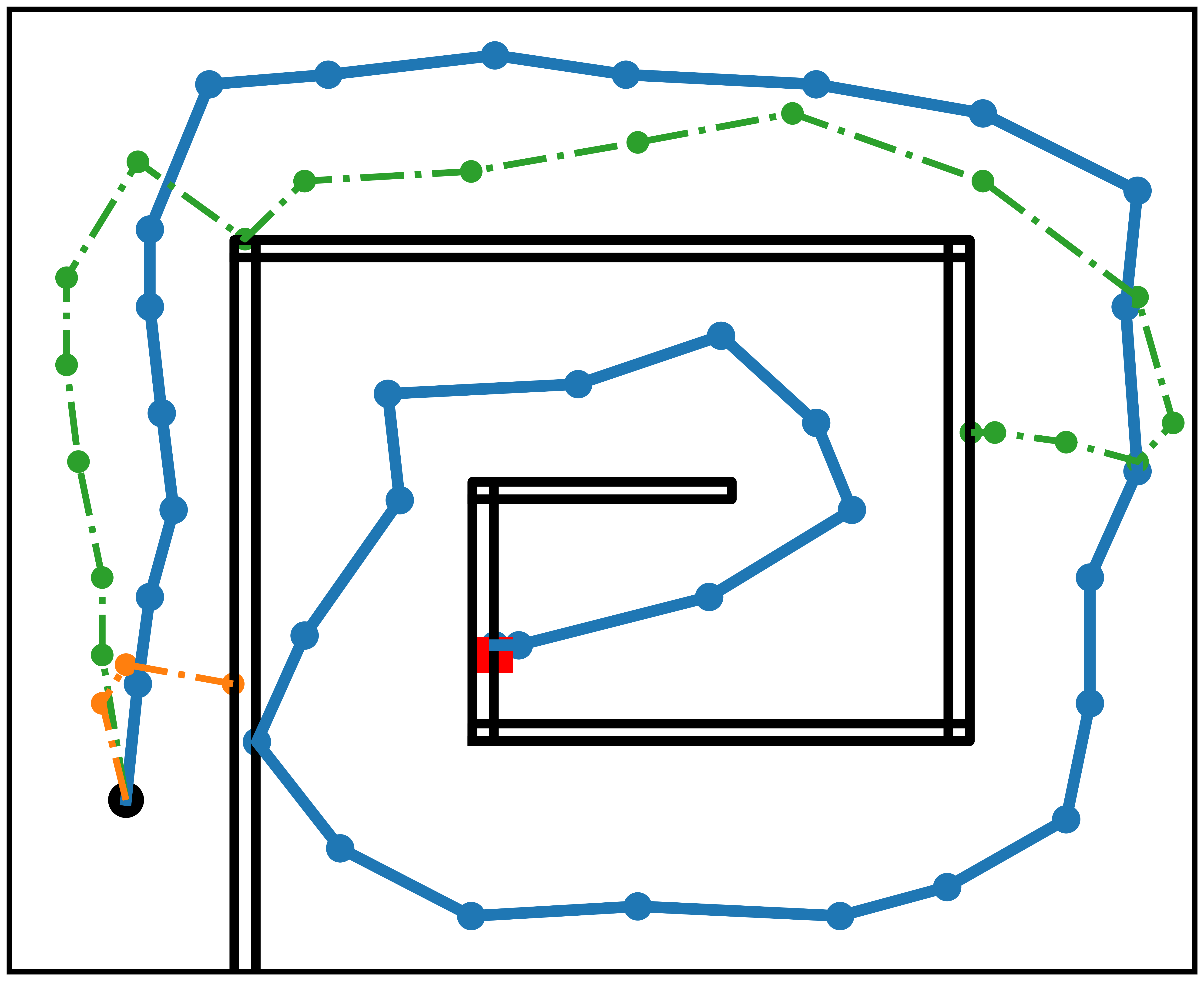}
    \caption{Planning Trajectories}
    \label{fig:maze_experiment_spiral_trajectories_pclast}
    \end{subfigure}%
\end{minipage}
\caption{
Clustering, Abstract-MDP, and Planning are shown for \emph{Maze-Spiral} environment. Details same as \cref{fig:maze_experiment_hallway}.}
\label{fig:maze_experiment_spiral}
\end{figure}

We now investigate the quality of learned latent representations by visualizing relationships created by them across true states. This is done qualitatively by clustering the learned representations of observations using \textit{k}-means. Distance-based planners use this relationship when traversing in latent space. In \cref{fig:maze_experiment_hallway_pclast,fig:maze_experiment_spiral_pclast}, we show clustering of \methodname representation of offline-observation datasets for \emph{Maze-Hallway} and \emph{Maze-Spiral} environment. We observe clusters having clear separation from the walls. This implies only states which are reachable from each other are clustered together. On the other hand, with ACRO representation in \cref{fig:maze_experiment_hallway_acro,fig:maze_experiment_spiral_acro}, we observe disjoint sets of states are categorized as single cluster such as in cluster-10 (orange) and cluster-15 (white) of \emph{Maze-Hallway} environment. Further, in some cases, we have clusters which span across walls such as cluster-14 (light-pink) and cluster-12 (dark-pink) in \emph{Maze-Spiral} environment. These disjoint sets of states violate a planner's state-reachability assumption, leading to infeasible plans.

\subsection{Multi-Level Abstraction and Hierarchical Planning}
\label{subsec:impact-multi-level-abstraction-planning}
In \cref{subsec:representation-goal-rl}, we found \methodname embedding improves goal-conditioned policy learning. However, reactive policies tend to generally have limitations for long-horizon planning. This encourages us to investigate the suitability of \methodname for \emph{$n$-level} state abstraction and hierarchical planning with \cref{alg:mpc} which holds promise for long-horizon planning. Abstractions for each level are generated using $k$-means with varying $k$ over the \methodname embedding as done in \cref{subsec:impact-state-abstraction}. We do not consider ACRO embedding due to poor clustering behavior shown in \cref{subsec:impact-state-abstraction}.

For simplicity, we begin by considering \emph{2}-level abstractions and refer to them as \textit{high} and \textit{low} levels. In \cref{fig:maze_experiment_hallway_pclast,fig:maze_experiment_spiral_pclast}, we show the learned high-level clusters. The corresponding transitions models between the abstract discrete states are shown in \cref{fig:maze_experiment_hallway_state_transitions,fig:maze_experiment_spiral_state_transitions}. Note that they match the true topology of the corresponding mazes. Using this discrete state representation, MPC is applied with the planner implemented in Algorithm~\ref{alg:mpc}.
The planned trajectories are shown in \cref{fig:maze_experiment_hallway_trajectories_pclast,fig:maze_experiment_spiral_trajectories_pclast} and agent's performance is reported in \cref{tab:expr_result} (last row). It suggests, \emph{$n$-Level planner} ($n=2$) generates feasible and shortest plans (blue line) in all cases.
As a baseline, we directly evaluate our \emph{low-level} planner with \methodname map ($\psi$) (green line) representation for cost minimization (\cref{eq:app_traj_opt}) which struggles due to long-horizon planning demand of the task and complex navigability of the environment and leads to suboptimal results. At the same time, we observe \emph{low-level} planner with ACRO representation (orange line) for cost minimization fails in all the cases. This is simply due to lack of neighborhood structure in ACRO representation.

\begin{figure}[tbh!]
    \centering
    \includegraphics[scale=0.52]{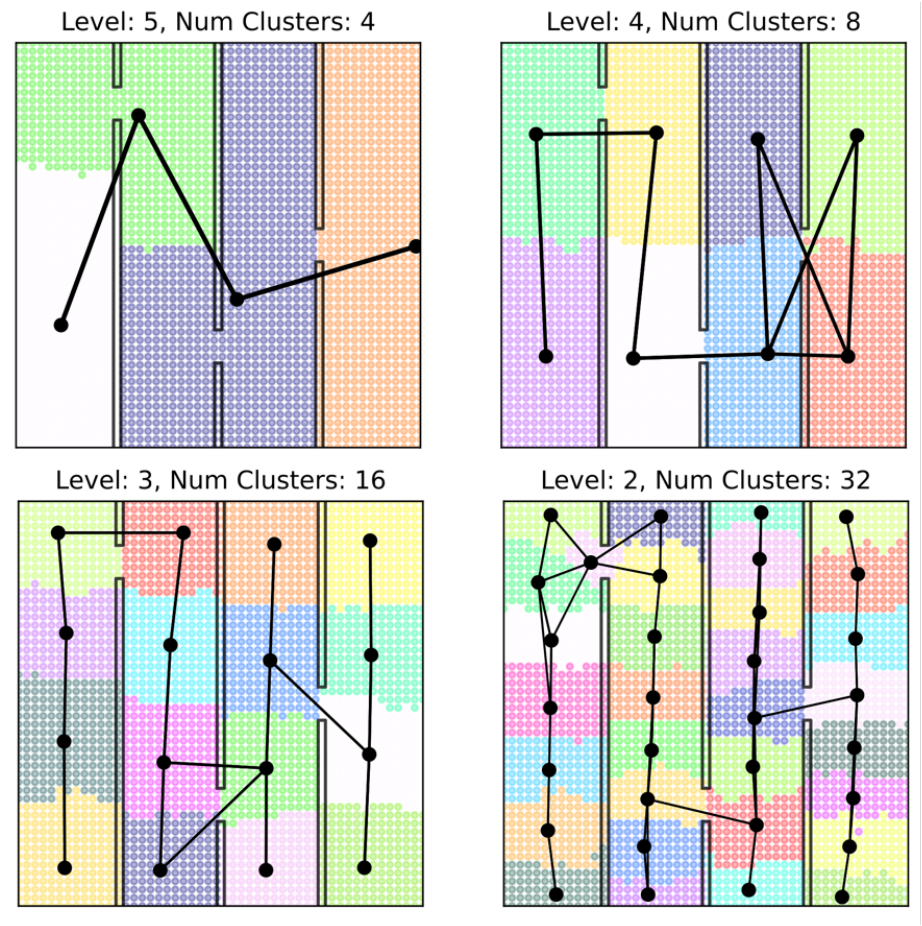}
    \vspace{-2mm}
    \caption{Visualization of hierarchical graphs in the \emph{Maze-Hallway} environment. At every level, num clusters is the $k$ used for clustering. The transition graph (in black) constructed from the cluster centers is superimposed over the environment. 
    }
    \label{fig:multi-level-abstraction-maze2d}
\end{figure}

\textbf{Increasing Abstraction Levels.} We investigate planning with multiple abstraction levels and consider $n \in \{2, 3, 4,5\}$. Performance scores for $n=5$ are reported in \cref{tab:expr_result} (bottom row). These abstractions help us create a hierarchy of graphs that describes the environment. In \cref{fig:multi-level-abstraction-maze2d}, we use $k=\{32,16,8,4\}$ for $n=\{2,3,4,5\}$ abstraction levels, respectively, and show graph-path for each abstraction-level for planning between two locations. This multi-level planning gives a significant boost to planning performance as compared to our model-free baselines. \emph{At the same time, we observe $3.8 \times$ computational time efficiency improvement in planning with $n=5$ (0.07 ms) as compared to $n=2$ (0.265 ms) abstraction levels}. However, no significant performance gains were observed by increasing levels. We assume this is due to the good quality of temporal abstraction at just $n=2$ levels which leads to the shortest plans and increasing the levels just helps to save on computation time. However, for more complex tasks, increasing the abstraction levels may further increase the quality of plans.

\subsection{Exogenous-Noise Offline RL Experiments}
\begin{figure}[tbh!]
    \centering
    \begin{subfigure}[t]{0.65\linewidth}
         \centering
         \includegraphics[width=\linewidth]{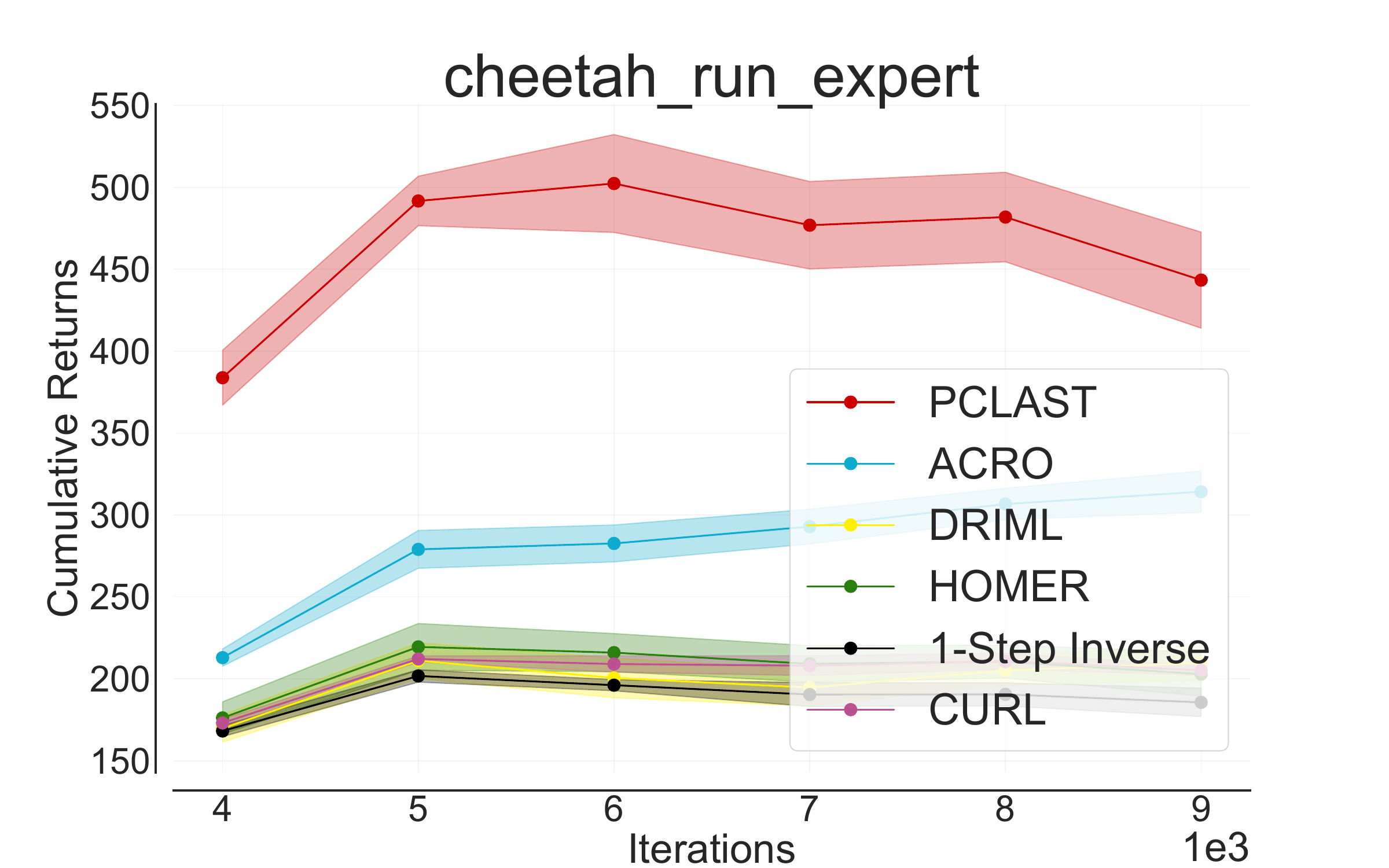}
         \label{fig:cheetah_expert}
    \end{subfigure}
    \begin{subfigure}[t]{0.65\linewidth}
         \centering
         \includegraphics[width=\linewidth]{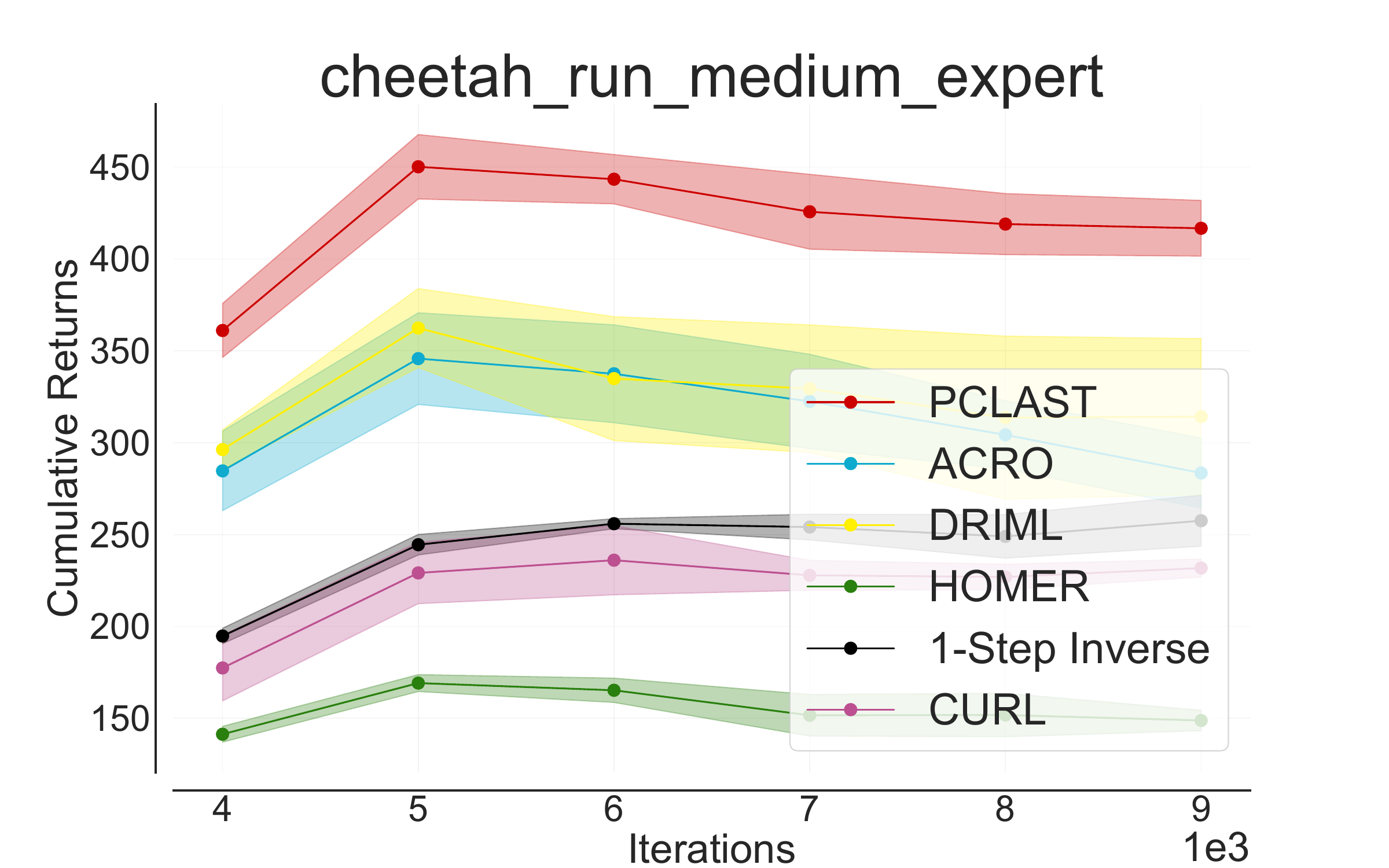}
         \label{fig:cheetah_medium_expert}
    \end{subfigure}
    \begin{subfigure}[t]{0.66\linewidth}
         \centering
         \includegraphics[width=\linewidth]{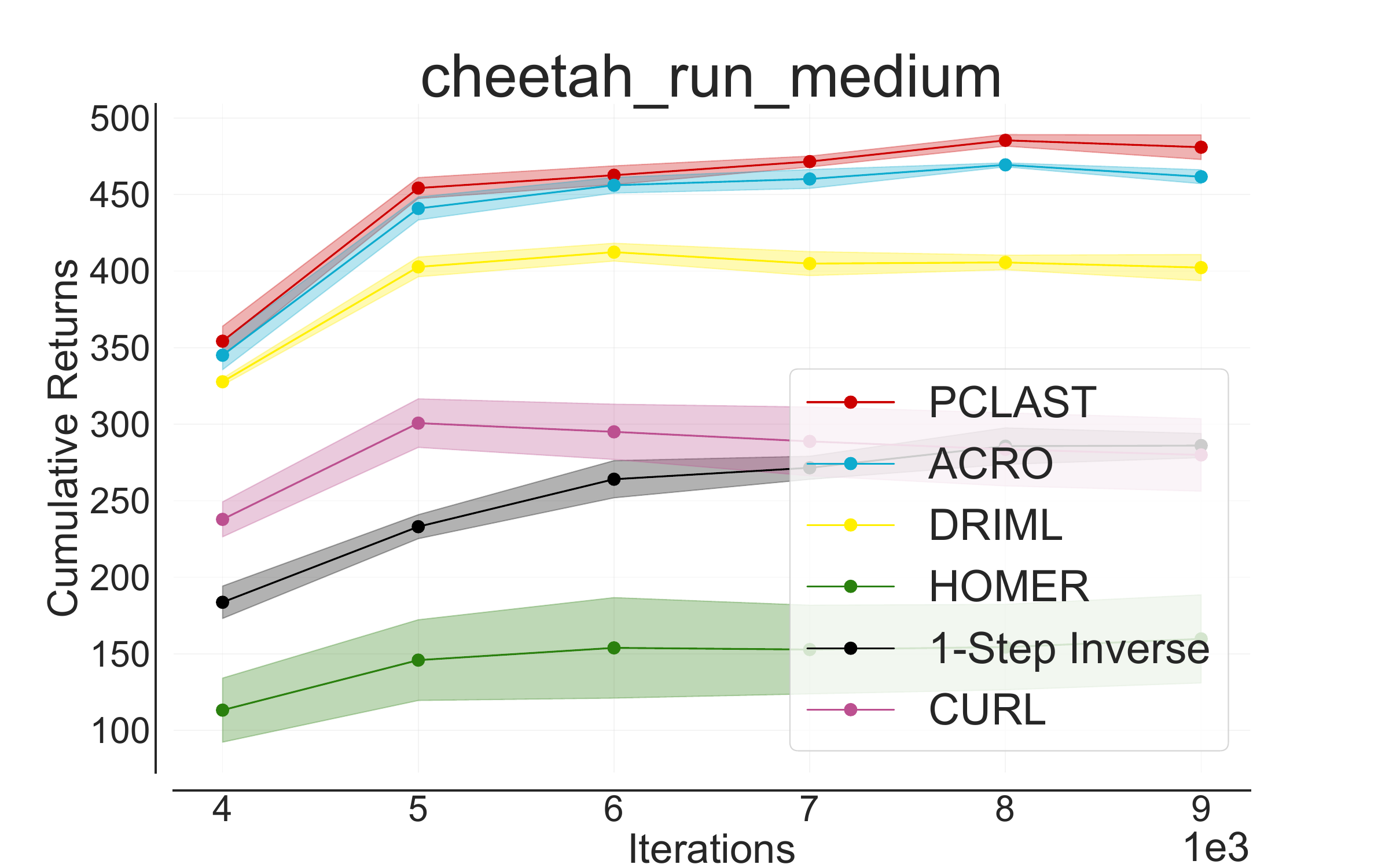}
         \label{fig:cheetah_medium}
    \end{subfigure}

    \caption{Comparisons of \methodname in \emph{Cheetah-Run} exogenous-noise environment with several other baselines.
    }
    \label{fig:offlineRL_results_main}
\end{figure}

Here, we evaluate \methodname exclusively on exogenous noised control environments described in \cref{subsec:environments}. We follow the same experiment setup as done by \citet{islam2022agent} and consider ACRO \cite{islam2022agent}, DRIML \cite{mazoure2020deep}, HOMER \cite{misra2020kinematic}, CURL \cite{laskin2020curl} and 1-step inverse model \cite{DBLP:conf/icml/PathakAED17} as our baselines. We share results for \emph{``Cheetah-Run"} with \emph{``expert, medium, and medium-expert"} dataset in \cref{fig:offlineRL_results_main}. It shows \methodname helps gain significant performance over the baselines \citep{islam2022agent}. Extended results for \emph{``Walker-Walk"} show similar performance trends as shown in \cref{fig:offlineRL_results} (Appendix). These results along with results in \cref{subsec:representation-goal-rl} suggest \emph{\methodname to be suitable for environments with non-linear dynamics} such as caused by presence of obstacles/walls.

\subsection{\methodname Ablations}
In \cref{subsec:architecture-implementation}, we detail our network architecture and hyperparameters. Here, we investigate critical hyperparameters in \methodname. These include: 1) the impact of $K_{max}$ on multi-step inverse dynamics, 2) cluster sizes in the $n$-Level planner, and 3) the importance of training a separate map $\psi$ instead of $\phi$ (ACRO) with contrastive loss (\cref{eq:lns_4}). These ablation studies are conducted in \emph{Maze} environments.

\begin{figure}[ht!]
    \centering
     \includegraphics[trim={0 0.3cm 6.5cm 0.27cm},clip,scale=0.6]{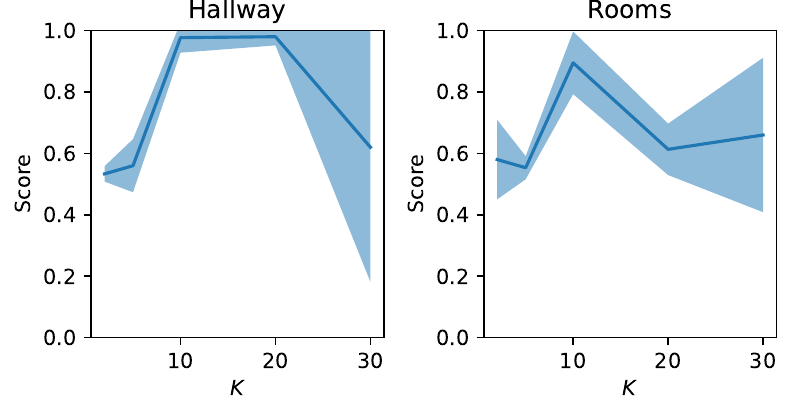}
    \caption{Effect of $K_{max}$ over agents performance in \emph{Maze} environments. The graph shows the normalized mean and standard deviation of scores over 3 seeds.}
    \label{fig:k_max_ablation}
\end{figure}

\textbf{$\boldsymbol{K_{max}}$.} 
In \cref{fig:k_max_ablation}, we see that agent performance improves with increasing $K_{max}$ but drops significantly when $K_{max}$ is too large. This is likely due to higher variance in action prediction in multi-step inverse dynamics ($f_{AC}$) for large values of $K_{max}$.

\begin{figure}[ht!]
    \centering
    \includegraphics[trim={0 0.3cm 6.5cm 0.27cm},clip,scale=0.6]{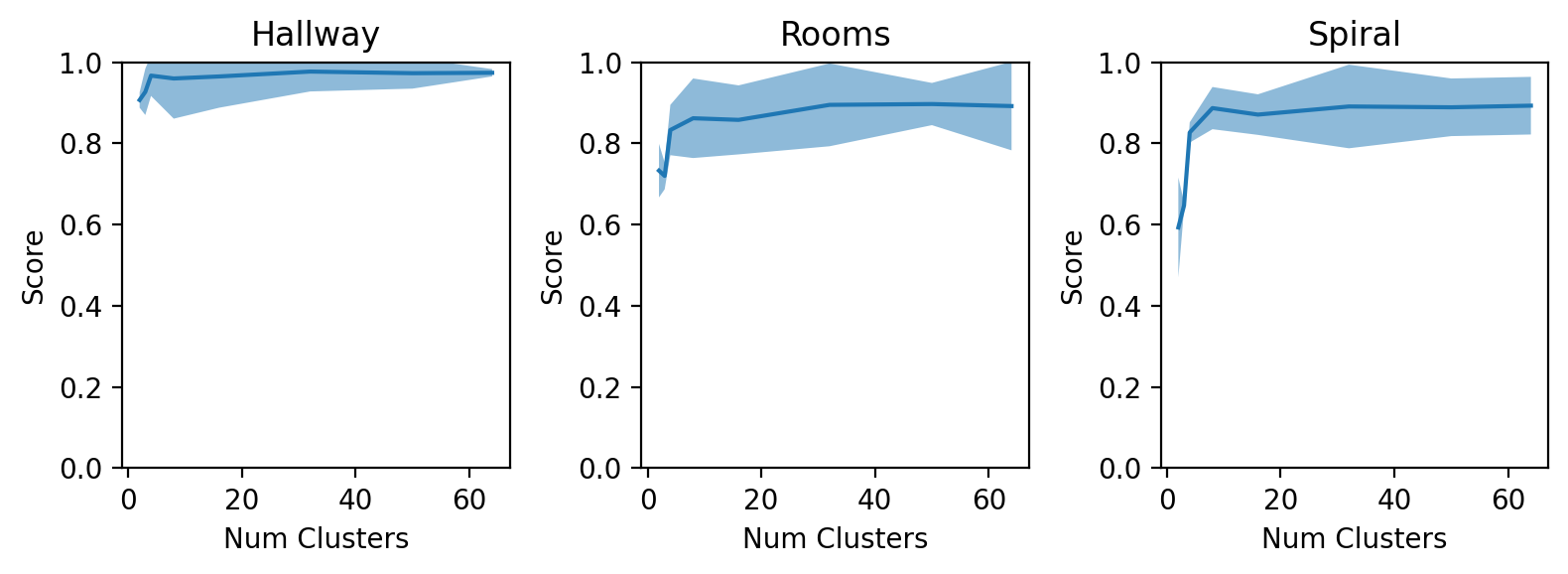}
    \caption{Effect of cluster sizes (C) for abstractions in $n$-Level planner over agent's performance in \emph{Maze} environments. The graph shows the normalized mean and standard deviation of scores over 3 seeds.}
    \label{fig:cluster_ablation}
\end{figure}

\textbf{Cluster-Size (C).} In \cref{fig:cluster_ablation}, we observe that increasing the number of clusters in $n$-Level (n=2) planner improves performance, as it makes clusters more robust to errors in the $\psi$ space, reducing planning errors.

\textbf{Contrastive Loss.} In \methodname, we learn a representation $\psi$ with the contrastive loss (\cref{eq:lns_4}). Here, we examine the effect on agent's performance if we use ACRO representation ($\phi$) trained via contrastive loss with $n$-Level planner instead of using $\psi$. The results are reported in \cref{tab:contrastive_loss_ablation}. As motivated in \cref{sec:map_learning}, \methodname tends to perform better than just applying contrastive loss on the ACRO ($\phi$).

\begin{table}[h!]
\caption{Comparing agent's performance when applying contrastive loss to ACRO ($\phi$) versus using the \methodname map $\psi$. Scores are reported over 3 seeds with $K_{max}=10$.}
\label{tab:contrastive_loss_ablation}
\begin{center}
\begin{small}
\begin{sc}
    \begin{tabular}{|c|c|c|}
        \hline
        \multirow{2}{*}{Env.} & \multicolumn{2}{c|}{Score} \\
        \cline{2-3}
         & $\phi$ + Contrastive Loss & \methodname \\
         \hline
         \hline
           Hallway &  89.3 $\pm$ 8.1 & \textbf{97.8 $\pm$ 4.9} \\
           Rooms &  79.3 $\pm$ 9.2 & \textbf{89.6 $\pm$ 10.2} \\
           Spiral &  81.3 $\pm$ 10 & \textbf{89.2 $\pm$ 10.3}\\
         \hline
    \end{tabular}
\end{sc}
\end{small}
\end{center}
\end{table}
\section{Summary}
\label{sec:summary}

Learning competent agents to plan in environments with complex sensory inputs, exogenous noise, non-linear dynamics, along with limited sample complexity requires learning compact latent representations which maintain state affordances. Our work introduces an approach  
that learns a representation via a multi-step inverse model and temporal contrastive loss objective. This makes the representation robust to exogenous noise as well as retains local neighborhood structure. Our diverse experiments suggest the learned representation is better suited for reactive policy learning, latent-space planning as well as multi-level abstraction for computationally efficient hierarchical planning.\looseness=-1
\section*{Impact Statement}
Decision making agents for real-world problems face challenges of rich high-dimensional observations embedded with significant exogenous elements. These noisy rich observations limit the usage of agent-training algorithms. It is addressed either by hand-crafting relevant feature extraction or controlling observable world elements such as in factories; both of which require additional human-engineering/effort and are still erroneous. Neural Networks have also been used to learn compact representations of the world to improve sample efficiency of agent-training algorithms. However, when planning in compact representation space, agents still tend to underperform. Further, agents are generally trained for a specific task and struggle to adapt to unseen similar tasks; which requires more data collection and training, increasing economic cost.

In our work, we build on task-agnostic representation methods which is robust to exogenous noise element of the observations; thereby increasing adaptability of trained agents in more realistic environments as well as reducing human-engineering effort to control the environment. Further, the learned representation maintains navigability relation to underlying true state space; easing the generalization and adaptability of agents to unseen interpolated states. On top of that, our framework enables efficient zero-shot planning for new tasks in the environment by decomposing the world into multiple levels of abstraction. This opens up the door to efficient and feasible solutions of complex problems with long horizon solutions. It also reduces human/computational effort spend in collecting data for new tasks as well as reduces economic/carbon costs for training new agents.

In terms of applications, we believe our approach would benefit any decision making agent which is deployed in real-world and/or requires quick adaptability to new tasks such as last-mile delivery robots, house robots, autonomous driving, and others. Our approach takes a step towards robustifying agents to real-world elements and can potentially help with safer interaction with the real-world.

\bibliography{references}
\bibliographystyle{icml2024}

\newpage
\appendix
\onecolumn
\section{Limitations}
\label{section:limitation}
\methodname representation and $n$-Level hierarchical planning enables us to do computationally efficient plans as well as achieves higher success rate in goal conditioned tasks. However, we observe \methodname is primarily effective in non-linear dynamic environments (such as with obstacles), and conventional methods (such as PPO) tend to have similar performance in linear environments.  When training \methodname, one needs domain knowledge to have good candidate of $K_{max}$ value, which is used for learning multi-step inverse dynamics. Further, when clustering the representation with \methodname map for the lowest level of hierarchy, one re-needs the domain knowledge to determine good number of clusters. In practice, we examine the environment to determine cluster count such that it leads to near-linear dynamics within the state-action space of a cluster at the lowest level. We also assume the presence of offline dataset with significant coverage of task-relevant state-action space. Though, our method just relies on the coverage and not on the quality of behavior policy; implying even a random policy data is sufficient for our method.

\section{Intuitive Argument: Learning Local Neighborhood Structure via Contrastive Learning}
\label{sec:contrastive}

In \cref{sec:map_learning}, we introduced a contrastive learning objective which allows us to learn the underlying local neighborhood structure in scalable and differentiable way (see equations~\eqref{eq:lns_1}-\eqref{eq:lns_4}). Here, we elaborate on the intuition that resulted in this objective. We show that the newly introduced temporal contrastive loss can be derived assuming a natural diffusion process on the underlying dynamics. 

Consider a simple discrete time multi-dimensional Brownian motion in $\bar{S}$ state-space. The conditional probability to observe state $\bar{s}'$ after stepping for $k$ steps from state $\bar{s}$, denoted as $\mathbb{P}_k(\bar{s}' |\bar{s})$, is given as~(see \citet{durrett2019probability}, Section 7):
\begin{align}
    \mathbb{P}_k(\bar{s}' | \bar{s}) \propto \exp\left( -\frac{\norm{\bar{s}-\bar{s}'}^2}{\sigma_0 k} \right). \nonumber
\end{align}
With this fact in mind, we can study the distribution of the contrastive learning process which outputs the tuple $(y,\bar{s},\bar{s}')$. 
A simple way to define this process is as follows: (1) sample $\bar{s}$ uniformly from $\bar{S}$, (2) After k-steps, set $\bar{s}'$ by sampling $y\sim \mathrm{Bernoulli}(0.5)$; if $y=1$, set $\bar{s}'  \sim \mathbb{P}_k(\cdot | \bar{s})$ otherwise set $\bar{s}'$ by uniformly sampling from $\bar{S}$.
We refer to this process as the \emph{Contrastive Learning(CL) generating process}. The following can be readily derived from these assumptions (see proof in Appendix~\ref{app:theory}).

 \begin{restatable}{proposition}{Firstprop}\label{prop:1}
   Assume the tuple $(y,\bar{s},\bar{s}')$ is sampled via the CL generating process described above. Then, $\PP_k(y=1 \mid \bar{s},\bar{s}') = \mathrm{sigmoid}(c-b \norm{\bar{s}-\bar{s}'}^2),$ where $ \mathrm{sigmoid}(x)=\exp(x)/(\exp(x)+1).$
 \end{restatable}

Although this generating process does not take into account the geometry of the underlying space and subtle intricacies of the environment, for small time scales this process can capture the dynamics to a reasonable degree. Further, the contrastive learning objective follows directly from Proposition~\ref{prop:1}. This objective can be interpreted as a log-likelihood learning procedure of the CL generating process.

\clearpage
\section{Bayes Solution of Contrastive Loss}\label{app:theory}





\Firstprop*
\begin{proof}
    The proof following by direct analysis of the conditional probability distribution together with the assumption of the CL generating process, i.e., the underlying Brownian motion.

\begin{equation*}
    \begin{split}
    \mathbb{P}_k(y = 1 | \bar{s}, \bar{s}') &= \frac{\mathbb{P}_k(\bar{s} | \bar{s}' , y=1) \mathbb{P}_k(y=1 | \bar{s})}{\mathbb{P}_k(\bar{s}'|\bar{s}, y=1) \mathbb{P}_k(y=1 | \bar{s}) + \mathbb{P}_k(\bar{s}'|\bar{s}, y=0) \mathbb{P}_k(y=0 | \bar{s})}  \qquad\qquad\qquad\qquad\qquad \because \text{Bayes' rule} \\
      &= \frac{\mathbb{P}_k(\bar{s} | \bar{s}' , y=1) }{\mathbb{P}_k(\bar{s}'|\bar{s}, y=1)  + \mathbb{P}_k(\bar{s}'|\bar{s}, y=0)} \qquad\qquad\qquad\quad \because \mathbb{P}_k(y=1 | \bar{s})=\mathbb{P}_k(y=0 | \bar{s})=0.5 \text{; Bernoulli}\\ 
     &= \frac{\mathbb{P}_k(\bar{s} | \bar{s}' , y=1) }{\mathbb{P}_k(\bar{s}'|\bar{s}, y=1)  + 1/\left| \mathcal{\bar{S}} \right| } \qquad\qquad\qquad\qquad\qquad\qquad\qquad\qquad\qquad\quad \because {\mathbb{P}_k(\bar{s}'|\bar{s}, y=0) \approx  1/\left| \mathcal{\bar{S}} \right|} \\
    &= \frac{C \exp\left( -\frac{\norm{\bar{s}-\bar{s}'}^2}{\sigma_0 k} \right) }{C \exp\left( \frac{-\norm{\bar{s}-\bar{s}'}^2}{\sigma_0 k} \right)  + 1/\left| \mathcal{\bar{S}} \right| } \qquad\qquad \because \text{Assuming brownian motion (where }  C \text{ is a positive constant)} \\
    &= \frac{ \exp\left(\log(C\left| \mathcal{\bar{S}} \right| ) -\frac{\norm{\bar{s}-\bar{s}'}^2}{\sigma_0 k} \right) }{\exp\left(\log(C\left| \mathcal{\bar{S}} \right| ) -\frac{\norm{\bar{s}-\bar{s}'}^2}{\sigma_0 k} \right)   +1 } \\
    &=\mathrm{sigmoid}(c-b \norm{\bar{s}-\bar{s}'}^2) \qquad\qquad\qquad\qquad\qquad\qquad\qquad  \text{ where } b=1/(\sigma_0k) \text{ and } c=\log(C \left| \mathcal{\bar{S}}\right|) \nonumber
    \end{split}
\end{equation*}

\end{proof}

\section{Experiments}
We discuss design of the considered environments as well as implementation details of our work.

\subsection{Maze-2D point mass}
\label{app:2dmaze}

\textbf{Environment Setup.} The state $s_t$ of the point-mass experiment is the 2D position of a point and the action $a_t$ is the position displacement, i.e., $s_{t+1}= s_t + a_t$. The action is bounded by $\lVert a_t \rVert_\infty \leq 0.2$. In the presence of obstacles, the point mass starting from $s_t$ is moved along the direction of $a_t$ until it collides with an obstacle. Further, we have two reward variants for each maze : 1) \textit{Dense-reward} and 2) \textit{Sparse-reward}. In the dense case, the agent receives a reward for the first time it crosses a particular distance threshold from the goal. Specifically, if $d_g$ is the distance to the goal, the agent receives a reward $r$ encouraging it to go closer to the goal. The thresholds for reaching the goal, as well as the corresponding reward values are given below.

\begin{align*}
    r &= 0.25 & \text{if this is the first time } d_g < 0.1 \\
     &= 0.5 & \text{if this is the first time } d_g < 0.05 \\
      &= 1 & \text{if } d_g < 0.03
\end{align*}

In the sparse setting, the agent receives a reward of 1 when it's within a distance of 0.03 to the goal state. For evaluation, we randomize the start and goal states from across the maze so as to test the agent's ability to reach diverse goals. 

As shown in \cref{fig:maze_experiment_hallway,fig:maze_experiment_spiral} in the main paper, we consider three environments with distinct layouts of obstacles. In each of these environments, a dataset of 500K samples is collected using random actions. An instance of the dataset has \textit{$<$ obs-image, state, action, next-state, next-obs-image $>$}, where \textit{state} and \textit{next-state} are the coordinates of the agent in the maze and represent the true environment state as the obstacles are fixed. We use the transaction data to train the latent dynamics and extract an abstract transaction model using k-means clustering over the latent states with the continuous actions discretized with a resolution of $0.01$ for identifying transitions of the latent states across different clusters. 

\begin{figure}
    \centering
    \includegraphics[width=0.5 \textwidth]{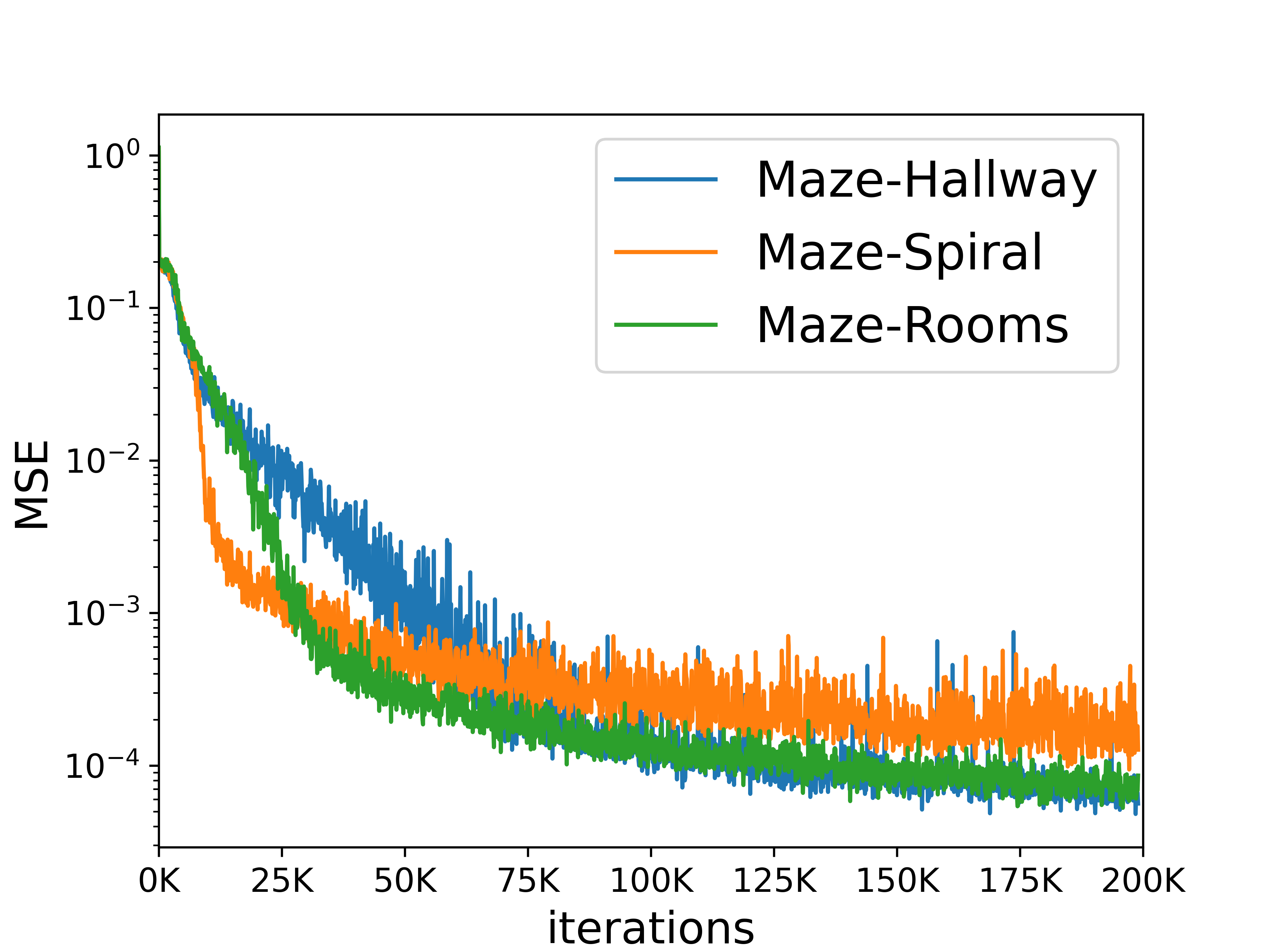}
    \caption{Mean square error of predicting the true state using the learned latent states during the training.}
    \label{fig:train_loss}
\end{figure}

\textbf{Latent Representation Quality.} Here we demonstrate the quality of the learned latent representation. Since, in the 2D point-mass experiment, we have access to the true state of the environment, we train a feed-forward network  to predict the true state from the learned latent state. In \cref{fig:train_loss}, we show the regression error of predicting the true state from the latent state. A significant low error implies that the learned latent state captures information about the true state. 

\textbf{Transition Model Generation.} In constructing the transition models between the clusters of latent states, we filter out the infrequent transitions to avoid giving hard-to-reach goals to the low-level planner. For the cluster $c_i$, let $c_{i_1}, c_{i_2}, \cdots, c_{i_N}$ denote the $N$ clusters such that there is at least a state transaction from $c_i$ to $c_{i_j}$ in the collected samples. This means it is feasible to move the point mass from $c_i$ to $c_{i_j}$. Let $0 < p_{i_j} \leq 1$ for $j = 1, \cdots, N$ denote the ratio of the number of transactions from $c_i$ to $c_{i_j}$ to the total number of outward transactions from $c_i$. We observe that if $p_{i_j}$ is small, the following issues may happen: (i) The transition from $c_i$ to $c_{i_j}$ is caused by the clustering errors and does not give a feasible transaction of the agent in practice. (ii) Even if such a transition is feasible, it is difficult for the low-level planner to find such a path as indicated by the sparsity of such transitions in the collected dataset. Therefore, in generating the transition models, we add the edge $c_i \rightarrow c_{i_j}$ to the graph only when $p_{i_j}$ is large enough. Without loss of generality, assume that $p_{i_j}$ for $j = 1, \cdots, N$ have been arranged in descending order. Motivated by the nucleus sampling \cite{holtzman2019curious}, we choose $N^* = \arg \min_{k} \sum_{j=1}^k p_{i_j} \geq 0.9$ and only add the edges $c_i \rightarrow c_{i_j}$ for $j = 1, \cdots, N^*$ to the graph. In this way, the spurious transitions are filtered out.

\subsection{Exogenous Noise Offline RL Experiments}
\label{appendix:exo_offline}
\begin{figure}[tbh]
    \centering
    \begin{subfigure}[t]{0.33\textwidth}
         \centering
         \includegraphics[width=\textwidth]{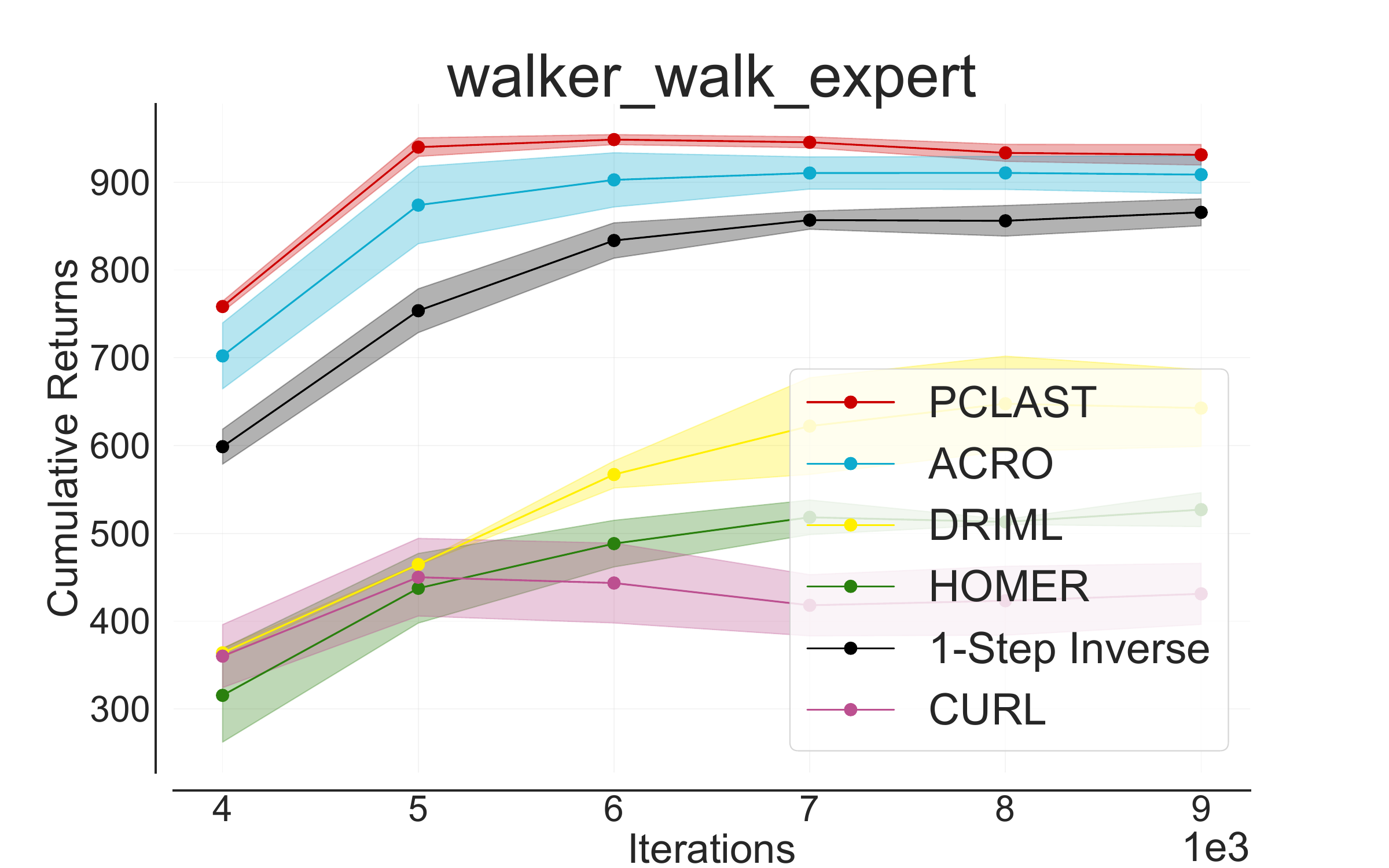}
         \label{fig:walker_walk_expert}
    \end{subfigure}
    \begin{subfigure}[t]{0.33\textwidth}
         \centering
         \includegraphics[width=\textwidth]{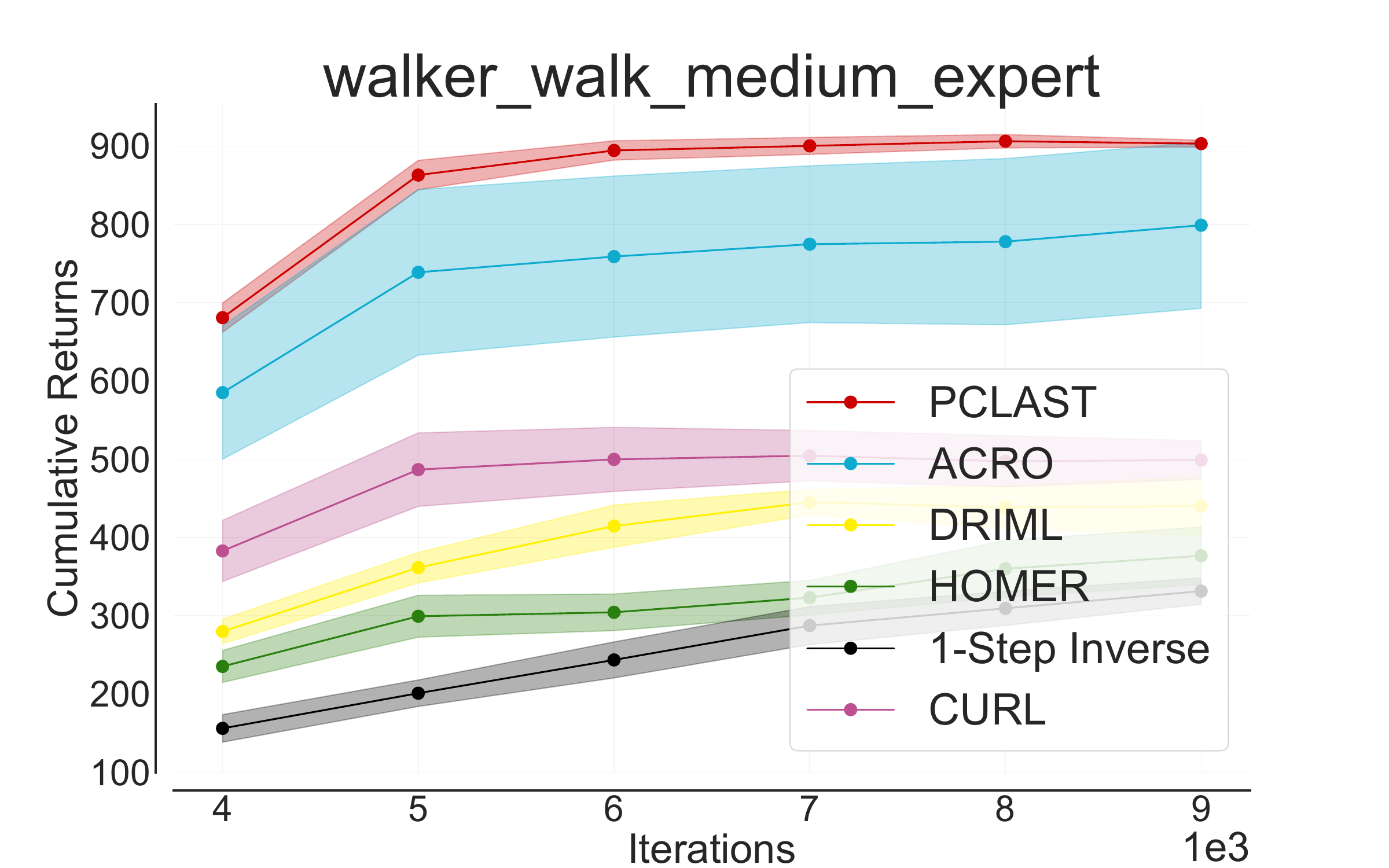}
         \label{fig:walker_walk_medium_expert}
    \end{subfigure}
    \begin{subfigure}[t]{0.30\textwidth}
         \centering
         \includegraphics[width=\textwidth]{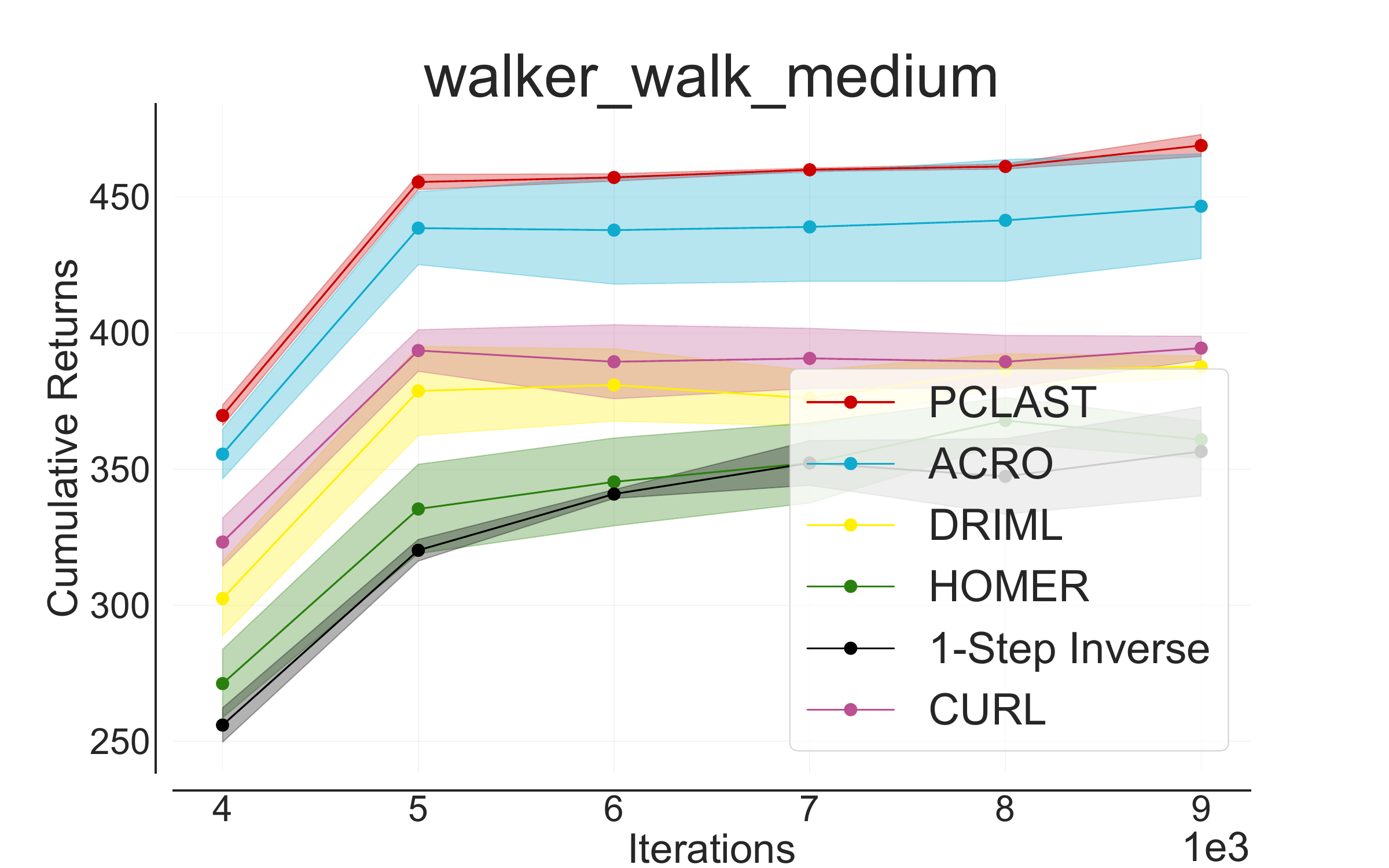}
         \label{fig:walker_walk_medium}
    \end{subfigure}\\
    \vspace{-3mm}
    \caption{Extended results of \methodname with several other baselines in \emph{Walker-Walk} task, following the  $4 \times 4$ exo-grid observation space offline RL setup from \cite{islam2022agent}.}
    \label{fig:offlineRL_results}
\end{figure}

We add exogenous noise by sampling 3 observations from ``random" quality dataset and adding them around the main observation as shown in Figure \ref{fig:offlineRL_observations_main}. This creates a  $4\times4$ exogenous observation from the offline dataset. This is same setup as from \cite{islam2022agent}. In our experiments, we have the controllable environment in one corner of the grid, and 3 other uncontrollable environments, taken from a random dataset, placed randomly in the $4 \times 4$ grid. Figure \ref{fig:offlineRL_results} shows additional results in the offline RL experimental setup.

\begin{figure}
    \centering
    \includegraphics[scale=0.3]{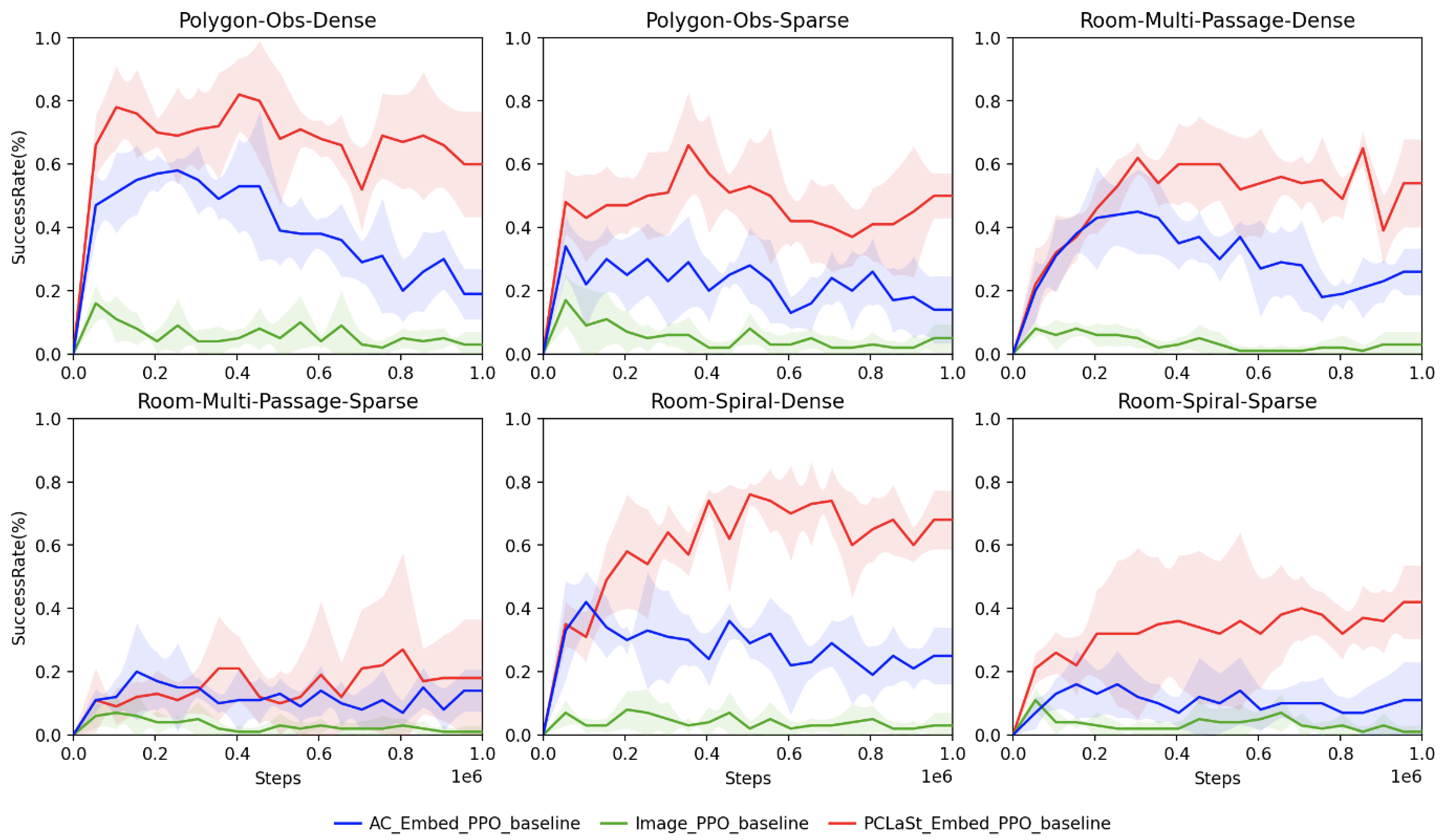}
    \caption{Performance comparison of PPO with ACRO and \methodname Embedding. The graph shows mean and standard deviation over 3 seeds.}
    \label{fig:ppo-comparison}
\end{figure}

\clearpage
\subsection{Network Architecture and Implementation Detail}
\label{subsec:architecture-implementation}

\methodname comprises of an encoder ($\phi$), multi-step inverse dynamics ($f_{AC}$), \methodname map  ($\psi$) and one-step forward dynamics ($\delta$). These networks are trained using Adam \cite{kingma2014adam} optimizer with learning rate of $1e^{-3}$ over batches of size 512. Unless specified otherwise, we sample transitions with $K_{max}=10$.  In the following, we discuss each of the networks. 

\begin{itemize}
\item \textbf{Encoder.} It comprises of 2 layer MLP-Mixer Network \cite{tolstikhin2021mlp} with an image patch size of 10, followed by a layer of BatchNorm \cite{ioffe2015batch} and GroupNorm \cite{wu2018group}. The output of GroupNorm is passed through a 2 layer network with intermediate Leaky ReLU activation. We use the same encoder output($\hat{s}$) dimension of 256 for all our experiments.

\item \textbf{Multi-Step Inverse Dynamics.}. It operates over a batch of $<\hat{s}_t,\hat{s}_{t+k},k>$. We use an Embedding layer to transform scalar $k$ into an embedding of size 45 and concatenate it with $\hat{s}_t$ and $\hat{s}_{t+k}$ before feeding the network. The inverse dynamics network is a multi-layer network with the first linear layer of size 256. This is followed by a 2-layer residual unit with in-between Gelu activation. Finally, we have two output head layers. The first outputs deterministic continuous actions and is trained with Mean square error loss. The second layer outputs categorical distribution over continuous values by binning the continuous action space and is trained via cross entropy loss. We divide the action-space of considered environments into 20 bins. During backprop, mean-square loss and categorical loss are scaled by 10 and 0.01 respectively. 

\item \textbf{\methodname map.} It's a 3-layer fully-connected network with Leaky ReLU as intermediate activations and hidden layers of size 512. For training, we sample positive examples using $d_m=2$, and negative samples are drawn randomly. Gradients from \methodname map are not propagated back to the encoder.

\item \textbf{Forward dynamics.} It's 4 layer fully-connected network with 512 as hidden layer size. It outputs a Gaussian distribution over the next state prediction. Again, gradients from forward dynamics are not propagated back to the encoder network.
\end{itemize}

\section{Multi-Level Planner Implementation Details}
\label{app:level}

\subsection{High-level planner}
\label{subsec:appendix-high-level-planner}
Given the current latent state $\hat{s}_t$ and the target latent state $\hat{s}^*$, the high-level planner aims to find an intermediate waypoint $\tilde{\hat{s}}$ such that the low-level planner can effectively track $\tilde{\hat{s}}$. The search for the waypoint is based on the discrete abstraction of the environment which is given by the graph $\mathcal{G}$ as described in Section~\ref{sec:learn_for_planning}. We denote $\phi_d(\cdot)$ as the node (or cluster) membership function for the graph abstraction $\mathcal{G}$ such that $\phi_d(\psi(\hat{s}))$ returns the node in $\mathcal{G}$ for any latent state $\hat{s} \in \mathcal{S}$. The high-level planner is outlined in Algorithm~\ref{alg:high_level}

\begin{algorithm}[tbh!]
\caption{High-level planner}
\label{alg:high_level}
\begin{algorithmic}[1]
\REQUIRE~~\\
Current latent state $\hat{s}_t$\\
target latent state $\hat{s}^*$\\
\methodname map $\psi(.)$\\
graph abstraction $\mathcal{G}$\\
\COMMENT{The node membership function $\phi_d(\cdot)$ is given by $\mathcal{G}$.}
\ENSURE{Waypoint $\tilde{\hat{s}}$}

\STATE Find the discrete latent states $c = \phi_d(\psi(\hat{s}))$ and $c^* = \phi_d(\psi(\hat{s}^*))$. \\
\IF{$c \neq c^*$}
    \STATE $\tilde{c} =$ Dijkstra($c, c^*, \mathcal{G}$)\\
    \COMMENT{$\tilde{c}$ is the next node on the shortest path from discrete states $c$ to $c^*$.}
    \STATE $\tilde{\hat{s}}$ = center of cluster $\tilde{c}$.
    \COMMENT{ $\tilde{\hat{s}} \in S$}
\ELSE
    \STATE $\tilde{\hat{s}} = \hat{s}^*, \tilde{c}= c^*$
\ENDIF
\end{algorithmic}
\end{algorithm}

\subsection{Low-level planner}
\label{subsec:appendix-low-level-planner}
Note that the latent forward dynamics is given by $\hat{s}_{t+1} = \delta(\hat{s}_t, a_t)$. Without loss of generality, we consider $\hat{s}_0$ as the current latent state and $\hat{s}^*$ as the target latent state. Given a horizon $T \geq 1$, our low-level planner generates actions $\{a_t\}_{t=0}^{T-1}$ that drive the latent state $\hat{s}_t$ to reach $\hat{s}^*$ by solving a trajectory optimization problem using \methodname map $\psi$.

\begin{equation}
\label{eq:app_traj_opt}
\begin{aligned}
& \underset{a_0, \cdots, a_{T-1}}{\text{minimize}}  \quad \sum_{t=0}^{T} \lVert \psi(\hat{s}_t) - \psi(\hat{s}^*) \rVert^2 \\
& \text{subject to} \quad \hat{s}_{t+1} = \delta(\hat{s}_t, a_t), \ t = 0, \cdots, T-1.
\end{aligned}
\end{equation}
In this work, we apply CEM to solve the low-level planning problem~\eqref{eq:app_traj_opt}. CEM has been successfully applied in model-based reinforcement learning~\citep{finn2017deep,wang2019exploring,hafner2019learning} and is outlined in Algorithm~\cref{alg:cem}. In the CEM, the actions $\{a_i\}_{i=0}^{T-1}$ are drawn from a multivariate Gaussian distribution whose parameters (mean and covariance matrix) are updated iteratively to approximate the optimal distribution of actions.

\begin{algorithm}[tbh!]
\caption{Cross-Entroy method}
\label{alg:cem}
\begin{algorithmic}[1]
\REQUIRE~~\\
Number of iteration $N$\\
number of samples $M$ each iteration\\
parameter $K$\\
\ENSURE Action sequence $\{a_i^*\}_{i=0}^{T-1}$
\STATE Initialize a multivariate Gaussian distribution $\mathcal{N}(\mu^{(0)}, \Sigma^{(0)})$ with mean $\mu^{(0)} = \mathbf{0}$ and covariance matrix $\Sigma^{(0)} = I$. \\
\FOR{$j = 0, \cdots, N-1$}
\STATE Sample $M$ action sequences $\{a_i^{(m)} \}_{i=1}^{T-1}$ from $\mathcal{N}(\mu^{(j)}, \Sigma^{(j)})$ for $m = 1, \cdots, M$. \\
\STATE For each action sequence $\{a_i^{(m)} \}_{i=1}^{T-1}$, evaluate the rendered cost $J^{(m)}$ of Problem~\eqref{eq:app_traj_opt}.\\
\STATE Select the $K$ smallest costs from $\{ J^{(m)} \}_{m=1}^{M}$ and the corresponding action sequences $\{a_i^{*, (k)} \}_{i=1}^{T-1}$ for $k =1, \cdots, K$. \\
\STATE Update $\mu^{(j+1)}$ and $\Sigma^{(j+1)}$ as the mean and covariance of $\{a_i^{*, (k)} \}_{i=1}^{T-1}$, $k=1, \cdots, K$.
\ENDFOR
\STATE Sample the action sequence $\{a_i^*\}_{i=0}^{T-1}$ from $\mathcal{N}(\mu^{(N)}, \Sigma^{(N)})$ as the output. 
\end{algorithmic}
\end{algorithm}

\begin{figure}[tbh!]
    \centering
    \begin{subfigure}[t]{0.2\textwidth}
         \centering
         \includegraphics[width=\textwidth]{figures/planning_methods/hallway_3_lines.png}
         \label{fig:hallway_cem}
         \caption{Cross-Entropy Method}
    \end{subfigure}
    \begin{subfigure}[t]{0.2\textwidth}
         \centering
         \includegraphics[width=\textwidth]{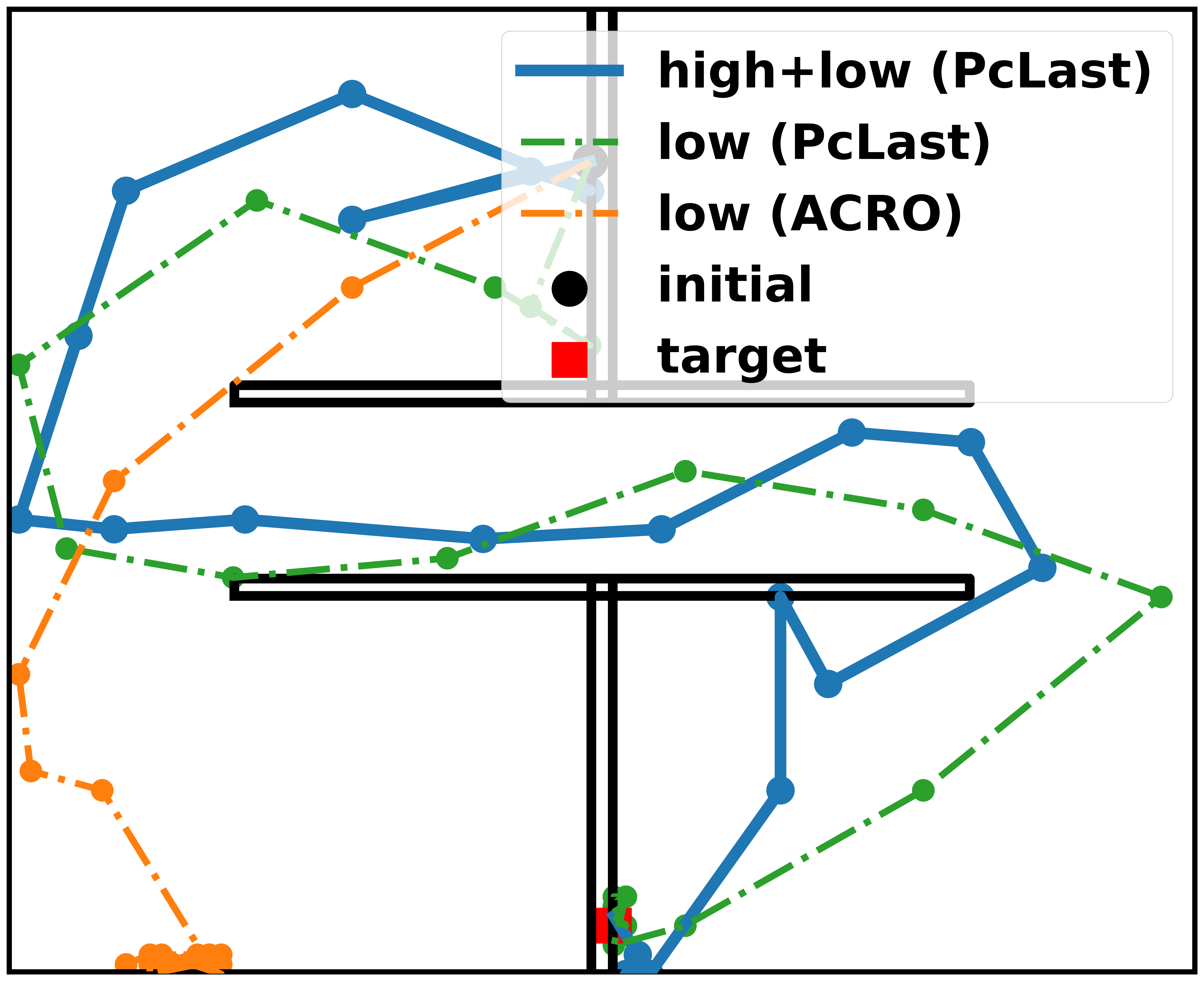}
         \label{fig:hallway_gd}
         \caption{Adam~\citep{kingma2014adam}}
    \end{subfigure}
    \caption{Comparison of the CEM (Left) and Adam (Right) methods in solving the low-level planning problem~\eqref{eq:app_traj_opt} in the Maze-Hallway environment. $16$ clusters are used for the high-level planner. It shows that the inherent complexity of problem~\eqref{eq:app_traj_opt} poses significant challenges for both low-level planners. A high-level planner shown in Algorithm~\ref{alg:high_level} is a necessary enabler of goal reaching. }
    \label{fig:low_planner_comparison}
\end{figure}

\paragraph{Low-level planners comparison.}
In \cref{fig:low_planner_comparison}, we evaluate the performances of the hierarchical planner and the low-level planner with two different methods, namely CEM and a first-order gradient descent-based method Adam~\citep{kingma2014adam}, to solve problem~\eqref{eq:app_traj_opt}. In both cases, the hierarchical planner reaches the goal while solely using the low-level planner fails. This means directly using the ACRO representation $\hat{s}$ for planning faces significant difficulty. \cref{fig:low_planner_comparison} shows that the high-level planner (Algorithm~\ref{alg:high_level}) derived on $\psi(\hat{s})$ is a key enabler of successful planning in the latent space and over a long horizon. 

\end{document}